%% file: main.tex
\documentclass{article}

\usepackage{iclr2027_conference,times}
\input{math_commands.tex}

\usepackage{latexsym}
\usepackage[T1]{fontenc}
\usepackage[utf8]{inputenc}
\usepackage{microtype}
\usepackage{inconsolata}
\usepackage{graphicx}
\usepackage{float}
\usepackage{placeins}
\usepackage{booktabs}
\usepackage{tabularx}
\usepackage{enumitem}
\usepackage{amsmath}
\usepackage{amssymb}
\usepackage{multirow}
\usepackage[table]{xcolor}
\usepackage{url}
\usepackage{pifont}
\usepackage{hyperref}

\newcommand{\method}{ReMind}
\newcommand{\gain}[1]{{\color{red}\scriptsize{+#1}}}

\title{Learning to Solve, Forgetting to Retain: Correct-Set Turnover in RLVR}

\iclrfinalcopy
\author{\parbox{\dimexpr\textwidth-2\tabcolsep\relax}{\centering
  Chuanyu Qin$^{1,2,*}$\quad Chenxu Yang$^{1,2,*}$\quad Qingyi Si$^{3,*}$\\
  Naibin Gu$^{1,2}$\quad Peng Fu$^{1,2,\dagger}$\quad Zheng Lin$^{1,2}$\\[0.5em]
  \normalfont\small
  $^1$Institute of Information Engineering, Chinese Academy of Sciences\\
  $^2$School of Cyber Security, University of Chinese Academy of Sciences\\
  $^3$JD.COM\\[0.3em]
  \texttt{qinchuanyu@iie.ac.cn}\quad\texttt{fupeng@iie.ac.cn}\\[0.3em]
  \footnotesize $^*$Equal contribution.\quad $^\dagger$Corresponding author.
}}
\hypersetup{
  hidelinks,
  pdftitle={Learning to Solve, Forgetting to Retain: Correct-Set Turnover in RLVR},
  pdfauthor={Chuanyu Qin, Chenxu Yang, Qingyi Si, Naibin Gu, Peng Fu, Zheng Lin}
}

\begin{document}
\raggedbottom
\maketitle
\lhead{Preprint}

\begin{abstract}
\input{sections/00_abstract}
\end{abstract}

\begin{figure}[H]
    \centering
    \includegraphics[width=\columnwidth]{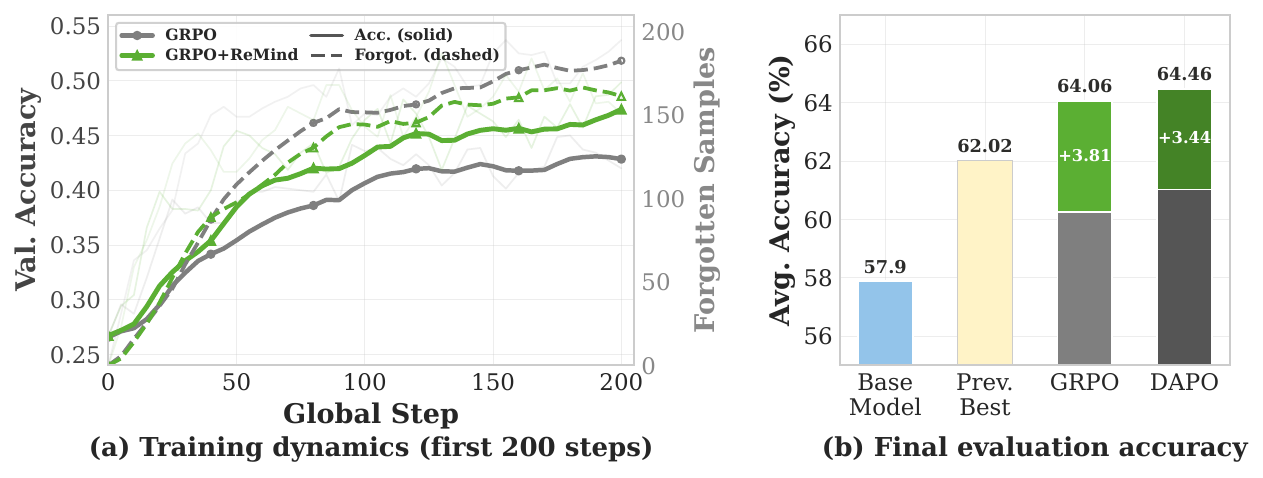}
    \caption{(a)~Training dynamics during the first 200 steps: validation accuracy (solid lines, left axis) and correct-set loss (dashed lines, right axis). GRPO accumulates forgetting steadily while accuracy plateaus; \method{} reduces forgetting and sustains accuracy gains under the same rollout budget. (b)~Average final accuracy across eight multimodal reasoning benchmarks. \method{} improves both GRPO and DAPO baselines.}
    \label{fig:intro-teaser}
\end{figure}

\clearpage
\begingroup
\setlength{\parskip}{3pt}
\input{sections/01_introduction}
\par
\endgroup
\input{sections/02_related_work}

\input{sections/03_preliminary}
\input{sections/04_method}
\input{sections/05_experiments}
\input{sections/06_conclusion}
\input{sections/08_ethics}

\setlength{\bibsep}{4pt plus 1pt minus 1pt}
\bibliography{references}
\bibliographystyle{iclr2027_conference}

\clearpage
\appendix
\input{sections/09_appendix}
\input{sections/07_limitations}

\end{document}

%% file: math_commands.tex
\usepackage{amsmath,amsfonts,bm}

\def\eqref#1{equation~\ref{#1}}

\def\1{\bm{1}}

\DeclareMathAlphabet{\mathsfit}{\encodingdefault}{\sfdefault}{m}{sl}
\SetMathAlphabet{\mathsfit}{bold}{\encodingdefault}{\sfdefault}{bx}{n}



%% file: sections/00_abstract.tex
Reinforcement learning with verifiable rewards (RLVR) improves the reasoning capabilities of large language models, yet aggregate accuracy gains can conceal a hidden cost: previously solved problems may regress as training proceeds.
We characterize this phenomenon as \emph{correct-set turnover}, capturing the coupled dynamics of solution acquisition and regression, and make retention an explicit training objective alongside acquisition.
In group-relative RLVR, mastered prompts provide no reward gradient when all rollouts are correct, yet can still drift under updates driven by other prompts.
This observation motivates a \emph{timely-review principle}: previously mastered prompts should be revisited before such regression accumulates.
We propose \textbf{ReMind}, a retention-aware review mechanism that tracks mastered prompts, periodically revisits them with fresh rollouts, and requeues those that regress.
Through pre-rollout batch replacement, ReMind preserves the per-step rollout budget.
We evaluate ReMind across 20 benchmarks spanning image-text, video, and text-only reasoning. On eight image-text benchmarks, ReMind improves average accuracy over GRPO and DAPO by \textbf{3.81} and \textbf{3.44} percentage points, respectively.

%% file: sections/01_introduction.tex
\section{Introduction}

Reinforcement learning with verifiable rewards (RLVR) has become a standard paradigm for improving the reasoning capabilities of large language models~\citep{guo2025deepseek, shao2024deepseekmath, yu2025dapo}. Headline accuracy gains, however, conceal a less visible cost: previously solvable problems can become unsolvable as training proceeds. RLVR has been shown to narrow the solution distribution, raising Pass@1 while reducing Pass@K~\citep{yue2025does, cui2025entropy}, while acquiring new reasoning skills can interfere with previously learned ones~\citep{hu2026emergentslowthinkingllms}. 
Such regression occurs within the same training distribution, distinct from the cross-task setting typically studied in continual learning.

Rather than viewing forgetting only as a side effect, we frame training progress symmetrically as two opposing flows: acquisition, where new problems become solvable, and regression, where previously solved problems are lost. Of particular interest are \emph{mastered} prompts, whose sampled rollouts are all correct at one point in training but may cease to be so after subsequent policy updates. We refer to these joint dynamics over the model's correct set as \emph{correct-set turnover}. This view makes retention an explicit training goal alongside acquisition, shifting attention from net accuracy gains to both learning new solutions and retaining previously solved ones.

Mastered prompts have an important property under group-relative RLVR: when all rollouts are correct, their group-relative advantage vanishes and they no longer contribute a reward gradient~\citep{shao2024deepseekmath, le2026rlzvp}. Policy updates are instead driven by prompts with mixed outcomes, yet these shared-policy updates can still cause mastered prompts to drift~\citep{liao2026mcpo}. Paradoxically, such drift can restore learnability: once a previously mastered prompt begins producing both correct and incorrect rollouts, it regains non-zero group-relative advantages and becomes informative for optimization again. This motivates allocating a small fraction of the rollout budget to revisiting mastered prompts, so that re-emerging corrective signals can be detected and used without substantially displacing fresh training. The remaining question is \emph{when} to spend this review budget. Inspired by the importance of review timing in spaced repetition~\citep{ebbinghaus1885memory, cepeda2006distributed}, we hypothesize that timely re-exposure should preserve mastered prompts more effectively than delayed review. Our controlled experiment in Figure~\ref{fig:repair_window} supports this hypothesis, showing larger retention gaps as review is delayed. Standard epoch-based RLVR pipelines~\citep{shao2024deepseekmath} revisit prompts only passively across epochs, with no explicit record of prior mastery.

We propose \textbf{\method{}},\footnote{Code is available at \url{https://github.com/cyuQ1n/Correct-Set-Turnover-in-RLVR}.} a lightweight retention-aware review mechanism that periodically reviews mastered samples to \textbf{remind} the model of previously solved solutions. \method{} maintains a queue of previously mastered prompts and periodically replaces a small fraction of fresh prompts with queued ones before rollout generation. Reviewed prompts pass through the standard rollout and policy-update pipeline: prompts that remain all-correct leave the queue, while those showing regression are re-enqueued for future review. This design preserves the per-step rollout budget while turning temporal changes in mastery into opportunities for corrective learning.

We evaluate \method{} across 20 benchmarks spanning image-text, video, and text-only mathematical reasoning with Qwen3-VL-8B-Instruct and Qwen2.5-7B-Math. On eight image-text benchmarks, \method{} improves GRPO, DAPO, and VCRL by 3.81, 3.44, and 1.54 percentage points on average, respectively, and outperforms replay and optimization-level baselines. These gains support compatibility with online filtering and curriculum learning. Training dynamics show lower correct-set loss with comparable acquisition coverage.

Our contributions are as follows.
\begin{itemize}[leftmargin=1.5em,topsep=2pt,itemsep=1pt,parsep=0pt,partopsep=0pt]
    \item We formalize \emph{correct-set turnover} in RLVR as the joint dynamics of acquisition and regression. Using a zero-overhead, graded mastery metric, we identify the \emph{timely-review principle}: revisiting mastered prompts earlier improves retention, with larger retention gaps as review is delayed.
    \item We propose \method{}, a retention-aware review mechanism that maintains a mastery-keyed review queue and uses pre-rollout replacement to detect and repair regression without increasing the rollout budget.
    \item \method{} improves late-stage accuracy and retention over standard RLVR baselines and replay methods across 20 benchmarks, demonstrating robust generalizability across different modalities, models, and algorithmic backbones.
\end{itemize}

%% file: sections/02_related_work.tex
\section{Related Work}
\paragraph{RL for LLMs.}
RLVR has become a standard post-training method for reasoning models~\citep{shao2024deepseekmath,guo2025deepseek}, with GRPO replacing the learned value model by group-relative comparison. Recent variants improve update stability and informativeness through dynamic sampling, clip-higher, and length-aware shaping (DAPO~\citep{yu2025dapo}), normalization bias correction~\citep{liu2025understanding}, and diagnostics on entropy and token-level objectives~\citep{cui2025entropy,dong2026probing}.
A related filtering and curriculum line~\citep{yu2025dapo,bae2025online,jiang2025vcrl} selects prompts whose rollouts are neither all correct nor all incorrect to keep useful gradients in the batch, treating uniformly correct prompts as low-value for the \emph{next update}.
RL-ZVP~\citep{le2026rlzvp} recovers learning signals from zero-variance groups, while MCPO~\citep{liao2026mcpo} stabilizes mastered prompts during optimization. \method{} instead treats mastery temporally, revisiting previously mastered prompts as training continues.

\paragraph{Replay and forgetting.}
Experience replay reuses past data for efficiency~\citep{lin1992experience,mnih2015human}, with prioritization based on estimated learning value~\citep{schaul2015prioritized}. Recent LLM RL methods replay successful or informative trajectories along various axes, including verified solutions~\citep{zhang2025rlep}, difficulty and entropy buckets~\citep{liang2025exgrpo}, response reuse and early-state replay~\citep{zhang2025ar3po,dou2025rrl}, and age-decayed priorities~\citep{ma2026freshper}, but none defines priority by whether a once-mastered prompt has regressed. This differs from continual learning, where forgetting arises from sequential exposure to disjoint tasks~\citep{kirkpatrick2017overcoming,rebuffi2017icarl}; we study regression within a single RL training distribution. Our framing is closest to example-forgetting analysis, where forgetting is a sample-level 1$\to$0 transition under single-evaluation labels~\citep{toneva2019empirical}. This binary view is natural in supervised settings, but in group-based RLVR each prompt is already evaluated through multiple rollouts; treating these as a graded mastery estimate exposes partial regression that single-evaluation definitions cannot register, which \method{} uses to detect and repair regression during review. Recent RLVR work also reports broader regression phenomena, including cross-task capability loss~\citep{phan2025beyond}, narrowed pass@$k$ distributions~\citep{yue2025does}, and collapse dynamics~\citep{hu2026emergentslowthinkingllms}, but does not turn sample-level mastery into an online training state for retention decisions.

%% file: sections/04_method.tex
\section{Correct-Set Turnover and \method{}}
\label{sec:method}

\subsection{Definition and Observation}
\label{sec:turnover:def}

Let $\mathcal{P}$ denote the training set.
For each sample $p\!\in\!\mathcal{P}$, we write
$q_t(p)=\Pr[\pi_{\theta_t}\text{ solves }p]$
for the success probability under the current policy $\pi_{\theta_t}$,
estimated with $K$ rollouts as
$\hat{q}_t(p)=\tfrac{1}{K}\sum_{i=1}^{K}\mathbf{1}[r_i\!=\!1]$.
A sample is \emph{mastered} at step $t$ if all rollouts are correct
($\hat{q}_t(p)=1$), yielding the mastered set
$M_t=\{p\in\mathcal{P}:\hat{q}_t(p)=1\}$.
Two properties of $\hat{q}_t(p)$ matter for what follows. It is \emph{graded}: a previously mastered sample at $\hat{q}_t = 12/16$ shows partial regression that a single correct/incorrect outcome cannot distinguish from full mastery, whereas the binary signal used in prior example-forgetting work~\citep{toneva2019empirical} collapses this regime. It is also \emph{free}: $\hat{q}_t(p)$ is the same per-prompt success rate GRPO computes for advantage estimation, so retention can be checked whenever a sample is included in the training batch without any auxiliary evaluation.

Comparing $M_t$ with a later $M_{t+\Delta}$ decomposes the net change in mastery into two opposing flows: \emph{acquisition}, where samples enter the mastered set ($p\!\notin\!M_t,\;p\!\in\!M_{t+\Delta}$), and \emph{regression}, where samples leave it ($p\!\in\!M_t,\;p\!\notin\!M_{t+\Delta}$). We call this continual flux \emph{correct-set turnover} and measure it with the retention rate:
\begin{align}
  \text{Retention}_{t}
  &= \frac{|M_t\cap M_{t+\Delta}|}{|M_t|}, \notag \\
  \text{Regression}_{t}
  &= 1 - \text{Retention}_{t}.
  \label{eq:retention}
\end{align}
Figure~\ref{fig:intro-teaser}(a) illustrates turnover on the validation set. Training Qwen3-VL-8B-Instruct with standard GRPO, we track validation accuracy alongside the count of samples that were once solved but are currently wrong (correct-set loss). The two signals reveal a telling asymmetry: accuracy rises and then plateaus after roughly 100 steps, yet correct-set loss climbs steadily throughout the same window, indicating that new acquisitions are increasingly offset by regressions on previously solved samples.

We also examine whether mastered training samples regress. Across 1,275 review records of previously mastered samples, 269 (about 21\%) contained both correct and incorrect rollouts, indicating partial regression from the earlier all-correct state.\footnote{These records were collected during a \method{} training run; see \S\ref{sec:method:overview} for the review procedure.}

\subsection{The Timely-Review Principle}
\label{sec:turnover:repair}

\begin{figure}[t]
  \centering
  \includegraphics[width=\columnwidth]{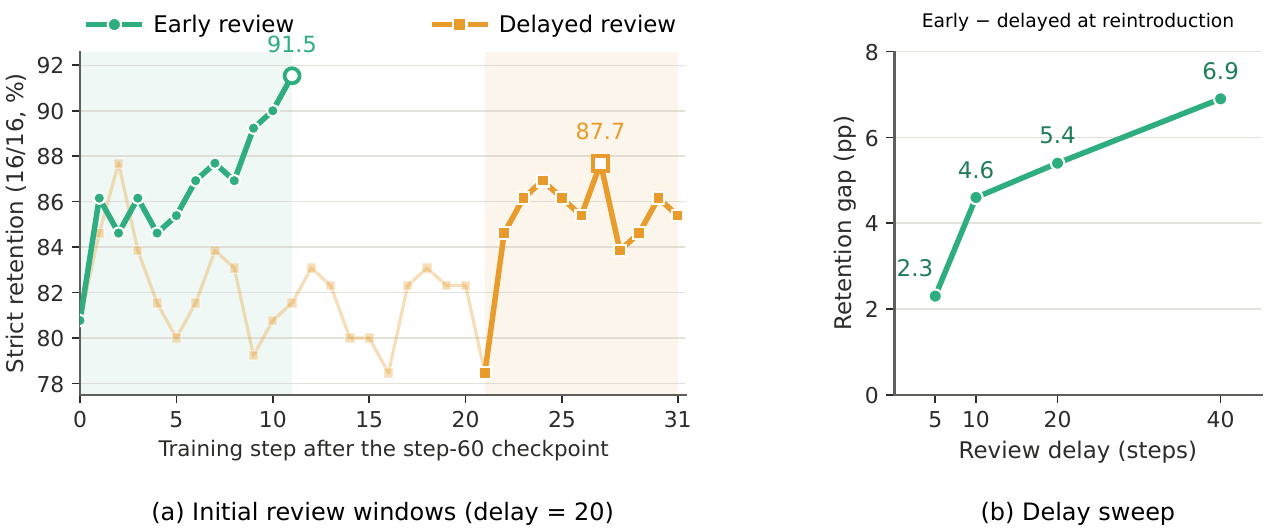}
  \caption{Strict retention (16/16 rollouts correct) under early and delayed review. (a)~Initial trajectories with a 20-step delay; shading highlights observation intervals. (b)~Retention gaps at reintroduction (early minus delayed, in percentage points) increase with review delay.}
  \label{fig:repair_window}
\end{figure}

To see why mastered prompts can regress during training, consider which samples contribute to GRPO's reward gradient. Under binary rewards, all-correct and all-incorrect rollout groups have zero group-relative advantages. Let $B_t$ denote the training batch and $g_t(p)$ the gradient of the GRPO reward loss for sample $p$. The batch reward gradient therefore comes only from groups with mixed outcomes:
\begin{equation}
  g_t^{\mathrm{RL}}
  = \frac{1}{|B_t|}
    \sum_{\substack{p\in B_t\\0<\hat{q}_t(p)<1}} g_t(p).
  \label{eq:reward_gradient}
\end{equation}
The resulting updates change the shared policy, including its behavior on mastered samples. They can therefore lower a mastered sample's success probability even while that sample provides no reward gradient of its own.

Partial regression creates an opportunity to counter these changes: when a previously mastered sample produces both correct and incorrect rollouts, its nonzero group-relative advantages provide a signal for reinforcing correct responses. Under epoch-based sampling without replacement, however, a sample is revisited only in a later epoch, leaving this corrective signal unused in a single-epoch run. This motivates \emph{review}: reintroducing previously mastered samples to generate fresh rollouts and use any resulting corrective signal. Samples that remain all-correct serve only to check retention. The next question is when to review these samples.

We compare immediate and delayed review of the same mastered prompts from a shared checkpoint. With a 20-step delay, retention is lower during most of the waiting period and improves after reintroduction (Figure~\ref{fig:repair_window}(a)). Across delays of 5, 10, 20, and 40 steps, the retention gap relative to immediate review at the same training step grows with delay (Figure~\ref{fig:repair_window}(b)).

These results establish a delay-sensitive retention effect: earlier review provides more timely opportunities to detect partial regression and reinforce correct responses. We refer to this as the \emph{timely-review principle}. The next subsection operationalizes this principle with \method{}.

\begin{figure*}[t]
  \centering
  \includegraphics[width=\textwidth]{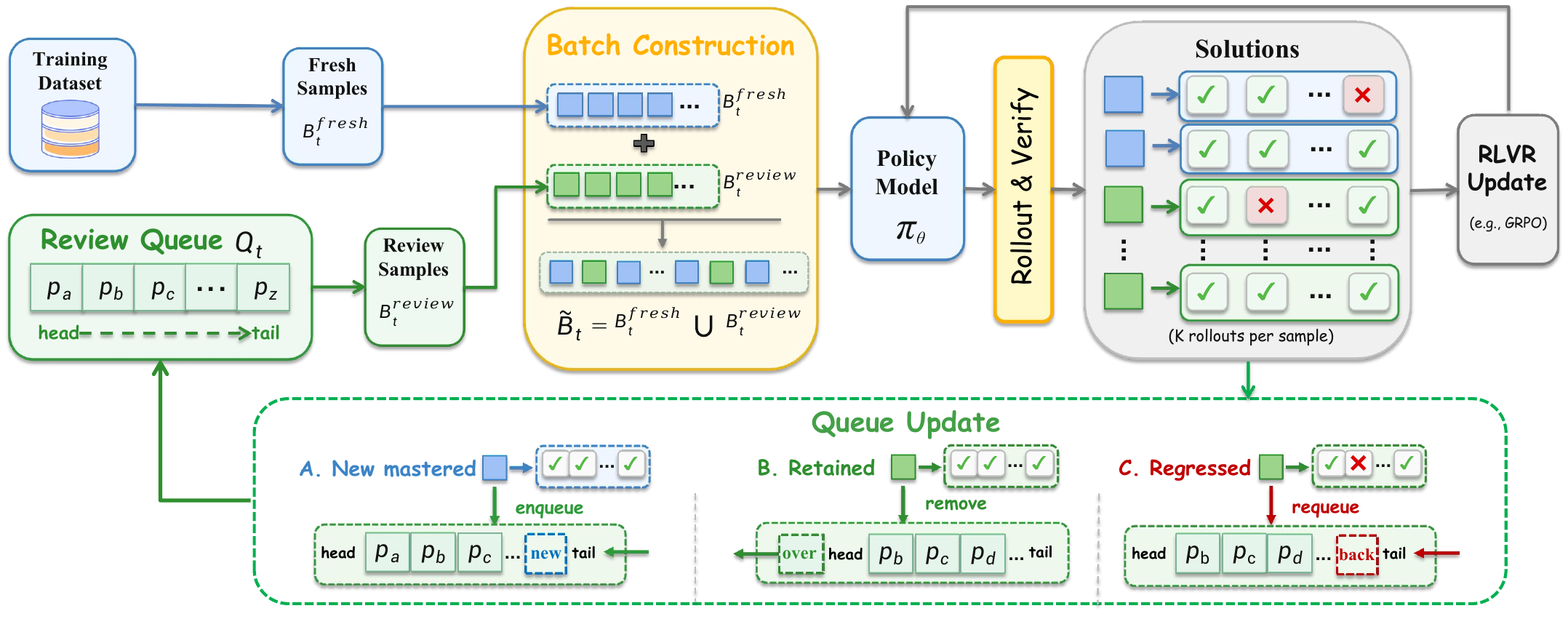}
  \caption{Overview of \method{}. The upper half combines fresh and queued review samples into a mixed batch $\widetilde{B}_t$ for rollout generation, verification, and the RLVR policy update. The lower half shows three queue-update cases: (A)~newly mastered samples are enqueued, (B)~retained samples are removed from the queue, and (C)~regressed samples are reinserted at the tail for future review.}
  \label{fig:method}
\end{figure*}

\subsection{\method{}}
\label{sec:method:overview}

\method{} implements the regular review motivated in \S\ref{sec:turnover:repair} through a queue that periodically reintroduces previously mastered samples into the training batch before rollout generation. This provides review opportunities within the current epoch. On review steps, queued samples replace a small fraction of the regular batch and pass through the same rollout, reward, and policy-update pipeline, keeping the per-step rollout budget unchanged. Figure~\ref{fig:method} illustrates the pipeline. Both mastery ($\hat{q}_t(p)=1$) and regression on previously mastered samples ($\hat{q}_t(p)<1$) are identified directly from the group rollouts GRPO already produces.

The method has three components: a FIFO review queue that stores mastered samples (\S\ref{sec:method:queue}), a batch-construction rule that mixes review and fresh samples on designated review steps (\S\ref{sec:method:batch}), and an update rule that routes regressed samples back into the queue for future repair (\S\ref{sec:method:update}).

\subsubsection{Review Queue}
\label{sec:method:queue}

Following the mastery definition in \S\ref{sec:turnover:def}, when a sample from the standard training stream satisfies $\hat{q}_t(p)=1$, \method{} appends it to the review queue with probability $p_{\mathrm{add}}$. The queue stores sample identifiers only; each subsequent review generates fresh rollouts from the current policy rather than replaying stored trajectories. FIFO ordering prioritizes samples that have waited longest since entering or re-entering the queue, providing a simple rule for allocating the review budget.

\subsubsection{Batch Construction}
\label{sec:method:batch}

Every $f$ steps is designated a \emph{review step}. On non-review steps, the batch is drawn entirely from the training sampler. On a review step, \method{} allocates a fraction $r$ of the batch to review samples drawn from the queue $Q_t$, forming a mixed batch:
\begin{align}
    \widetilde{B}_t &= B_t^{\mathrm{fresh}} \cup B_t^{\mathrm{review}},
    \notag \\
    |B_t^{\mathrm{review}}| &=
    \min\!\bigl(\lfloor r \cdot |\widetilde{B}_t| \rfloor,\;|Q_t|\bigr),
    \label{eq:batch}
\end{align}
where $B_t^{\mathrm{fresh}}$ contains samples from the standard stream and $B_t^{\mathrm{review}}$ contains samples dequeued from the head of $Q_t$. This mixed batch then proceeds through rollout, reward computation, and policy update without modification. The review ratio $r$ and period $f$ together control the fraction of total training devoted to retention review.

\subsubsection{Queue Update}
\label{sec:method:update}

After reward computation, the queue is updated according to three cases (lower panel of Figure~\ref{fig:method}). (1) Any sample from $B_t^{\mathrm{fresh}}$ that is mastered ($\hat{q}_t(p)=1$) and not already in the queue is enqueued with probability $p_{\mathrm{add}}$. (2) A review sample from $B_t^{\mathrm{review}}$ that retains full accuracy passes the current check and is not reinserted. We refer to removal after an all-correct review as graduation. (3) A review sample that regresses ($\hat{q}_t(p)<1$) is reinserted at the queue tail for future repair. This ensures regressed samples cycle back for additional review while samples passing the current check exit.

%% file: sections/05_experiments.tex
\section{Experiments}
\label{sec:experiments}

\subsection{Experimental Setups}
\label{sec:exp:setup}

\noindent \textbf{Datasets.}
We train on MMFineReason-123K~\citep{lin2026mmfinereasonclosingmultimodalreasoning}, a difficulty-filtered subset of MMFineReason-1.8M that retains only samples unsolved by Qwen3-VL-4B-Thinking~\citep{qwen3vl} across four rollouts. We evaluate on eight image-text reasoning benchmarks covering math reasoning (MathVista~\citep{lumathvista}, MathVision~\citep{wang2024measuring}, WeMath~\citep{qiao2025we}, MathVerse~\citep{zhang2024mathverse}) and general multimodal understanding (MMMU-Pro~\citep{yue2025mmmu}, MMBench~\citep{liu2024mmbench}, MM-Star~\citep{chen2024we}, ZeroBench~\citep{roberts2025zerobench}).

\noindent \textbf{Models and baselines.}
All image-text experiments use Qwen3-VL-8B-Instruct~\citep{qwen3vl} on a shared GRPO-style backbone.
We compare GRPO, DAPO, MCPO, RL-ZVP, ExGRPO, RLEP, RePO, and VCRL (Table~\ref{tab:main}).

\noindent \textbf{Implementation.}
We build on EasyVideoR1~\citep{qin2026easyvideor1easierrlvideo} with a batch size of 256 and $K\!=\!8$ rollouts per prompt unless otherwise specified.
For \method{}, we set the review ratio $r\!=\!0.10$, review period $f\!=\!5$, and enqueue probability $p_{\mathrm{add}}\!=\!0.25$, so that review samples constitute $\sim$2\% of total training prompts.
Full training hyperparameters and baseline descriptions are in Appendix~\ref{app:setup_details}.

\begin{table}[H]
\centering
\small
\caption{Main results on eight image-text reasoning benchmarks. Best and second-best scores are \textbf{bold} and \underline{underlined}; {\color{red}$\vartriangle$} rows show gains over the corresponding base method.}
\label{tab:main}
\resizebox{\textwidth}{!}{%
\begin{tabular}{l cccc c cccc c c}
\toprule
\multirow{2}{*}{\textbf{Method}}
& \multicolumn{5}{c}{\textbf{Math Reasoning}}
& \multicolumn{5}{c}{\textbf{General Multimodal}}
& \multirow{2}{*}{\textbf{Avg}} \\
\cmidrule(lr){2-6} \cmidrule(lr){7-11}
& Vista & Vision & WeMath & Verse & Avg
& MMMU-P & MMB & MM-S & ZeroB & Avg & \\
\midrule
Base Model
& 73.80 & 47.37 & 54.10 & 54.61 & 57.47
& 51.75 & 89.79 & 71.83 & 19.76 & 58.28 & 57.88 \\
\midrule
ExGRPO
& 77.30 & 55.46 & 62.67 & 56.89 & 63.08
& 55.49 & 90.44 & 72.00 & 19.01 & 59.24 & 61.16 \\
RLEP
& 78.50 & 54.23 & 62.48 & 58.91 & 63.53
& 55.38 & 90.45 & 72.27 & 19.61 & 59.43 & 61.48 \\
RePO
& 78.00 & 53.77 & 63.52 & 59.39 & 63.67
& 54.95 & 90.45 & 72.60 & 23.50 & 60.38 & 62.02 \\
\midrule
MCPO
& 76.50 & 53.75 & 62.10 & 59.12 & 62.87
& 54.35 & 90.07 & 71.13 & 23.35 & 59.73 & 61.30 \\
RL-ZVP
& 77.40 & 54.39 & 62.38 & 58.76 & 63.23
& 56.22 & 90.46 & 71.90 & 23.65 & 60.56 & 61.89 \\
\midrule
GRPO
& 76.20 & 48.82 & 56.57 & 59.52 & 60.28
& 55.78 & 90.29 & 72.20 & 22.60 & 60.22 & 60.25 \\
\rowcolor{cyan!10}
GRPO + \method{}
& 79.10 & 56.81 & 66.38 & \underline{61.52} & 65.95
& \underline{58.78} & 90.41 & 73.40 & \textbf{26.05} & 62.16 & 64.06 \\
\rowcolor{cyan!10}
{\color{red}$\vartriangle$} & \gain{2.90} & \gain{7.99} & \gain{9.81} & \gain{2.00} & \gain{5.67}
& \gain{3.00} & \gain{0.12} & \gain{1.20} & \gain{3.45} & \gain{1.94} & \gain{3.81} \\
\midrule
DAPO
& 77.60 & 52.78 & 57.43 & 59.55 & 61.84
& 56.85 & 90.53 & 72.77 & 20.66 & 60.20 & 61.02 \\
\rowcolor{cyan!10}
DAPO + \method{}
& \textbf{79.85} & \underline{57.32} & \underline{68.29} & 60.37 & \underline{66.46}
& 57.89 & \textbf{91.22} & \textbf{74.70} & \textbf{26.05} & \underline{62.47} & \underline{64.46} \\
\rowcolor{cyan!10}
{\color{red}$\vartriangle$} & \gain{2.25} & \gain{4.54} & \gain{10.86} & \gain{0.82} & \gain{4.62}
& \gain{1.04} & \gain{0.69} & \gain{1.93} & \gain{5.39} & \gain{2.27} & \gain{3.44} \\
\midrule
VCRL
& 77.75 & 56.81 & 64.76 & 61.31 & 65.16
& 58.25 & 90.53 & 72.93 & 25.00 & 61.68 & 63.42 \\
\rowcolor{cyan!10}
VCRL + \method{}
& \underline{79.45} & \textbf{58.50} & \textbf{68.48} & \textbf{63.00} & \textbf{67.36}
& \textbf{59.54} & \underline{90.60} & \underline{74.53} & \underline{25.60} & \textbf{62.57} & \textbf{64.96} \\
\rowcolor{cyan!10}
{\color{red}$\vartriangle$} & \gain{1.70} & \gain{1.69} & \gain{3.72} & \gain{1.69} & \gain{2.20}
& \gain{1.29} & \gain{0.07} & \gain{1.60} & \gain{0.60} & \gain{0.89} & \gain{1.54} \\
\bottomrule
\end{tabular}}
\end{table}

\clearpage
\subsection{Main Results}
\label{sec:exp:main}

\noindent\textbf{\method{} improves image-text reasoning across training algorithms.}
Starting from the same GRPO backbone, MCPO and RL-ZVP improve the overall average from 60.25 to 61.30 and 61.89, respectively (Table~\ref{tab:main}), supporting the value of explicitly handling high-accuracy and zero-variance prompts.
\method{} further raises the average to 64.06 (+3.81 over GRPO), outperforming MCPO and RL-ZVP by 2.76 and 2.17 points, respectively, as well as RePO, the strongest trajectory-replay baseline, by 2.04 points.\footnote{Three additional independent GRPO+\method{} training runs yield $64.18\pm0.03$ overall (mean $\pm$ sample standard deviation); Appendix~\ref{app:multi_seed} reports the per-seed results.}
Adding \method{} to DAPO and VCRL further improves their averages to 64.46 and 64.96, supporting its compatibility with dynamic sampling and curriculum learning.
Together, these results are consistent with \method{}'s temporal view that the utility of mastered prompts can re-emerge as training continues.

\noindent\textbf{The gains extend to video understanding and text math reasoning.}
Across six benchmarks in each domain (Table~\ref{tab:cross_modal}), \method{} improves the video average over GRPO by 3.21 points and exceeds ExGRPO on text math by 1.23 points.
Video-Holmes shows the largest video gain (+6.04); on MLVU, \method{} partially recovers GRPO's drop below the base model.
Benchmark details are in Appendix~\ref{app:video_benchmarks}.

\begin{table}[H]
\centering
\small
\setlength{\tabcolsep}{3pt}
\caption{Generalization of \method{} to video understanding and text math reasoning.
The base models are Qwen3-VL-8B-Instruct for video and Qwen2.5-7B-Math~\citep{yang2024qwen25mathtechnicalreportmathematical} for text math.
Best results within each block are in \textbf{bold}.}
\label{tab:cross_modal}
\begin{tabularx}{\textwidth}{@{}l*{7}{>{\raggedleft\arraybackslash}X}@{}}
\toprule
\multicolumn{8}{@{}l}{\textbf{(a) Video understanding}} \\
\addlinespace[2pt]
\textbf{Method} & MVB & Motion & MLVU & Holmes & LVR & VMQA & \textbf{Avg} \\
\midrule
Base Model
  & 69.50 & 56.10 & \textbf{78.40} & 45.30 & 77.30 & 47.80 & 62.40 \\
GRPO
  & 70.75 & 57.01 & 75.80 & 44.80 & 78.61 & 50.33 & 62.88 \\
\rowcolor{cyan!10}
\method{}
  & \textbf{71.22} & \textbf{61.44} & 77.37 & \textbf{50.84} & \textbf{82.49} & \textbf{53.19} & \textbf{66.09} \\
\midrule
\multicolumn{8}{@{}l}{\textbf{(b) Text math reasoning}} \\
\addlinespace[2pt]
\textbf{Method} & AIME24 & AIME25 & AMC & MATH & Minerva & Olympiad & \textbf{Avg} \\
\midrule
Base Model
  & 12.50 & 10.20 & 48.50 & 80.40 & 32.70 & 41.00 & 37.55 \\
SFT
  & 22.20 & 22.30 & 52.80 & 82.60 & \textbf{40.80} & 43.70 & 44.07 \\
SFT + RL
  & 25.80 & 23.10 & 62.70 & 87.20 & 39.70 & 50.40 & 48.15 \\
ExGRPO
  & 32.30 & 25.70 & 65.60 & 87.60 & 40.10 & 57.00 & 51.38 \\
\rowcolor{cyan!10}
\method{}
  & \textbf{33.85} & \textbf{26.25} & \textbf{67.85} & \textbf{90.80} & 38.97 & \textbf{57.93} & \textbf{52.61} \\
\bottomrule
\end{tabularx}
\end{table}

\subsection{Ablation Studies}
\label{sec:exp:ablation}
\label{sec:exp:review_strategies}

\begingroup
\setlength{\intextsep}{6pt}
\begin{figure}[H]
\begin{minipage}[t]{0.45\textwidth}
\vspace{0pt}
\centering
\setlength{\abovecaptionskip}{3pt}
\setlength{\belowcaptionskip}{0pt}
\includegraphics[width=\linewidth]{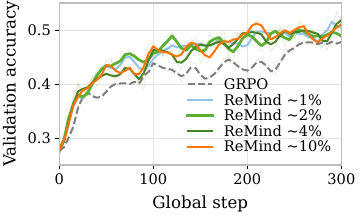}
\caption{Review budget sensitivity.}
\label{fig:ablation_curve}
\end{minipage}\hfill
\begin{minipage}[t]{0.52\textwidth}
\vspace{0pt}

\textbf{ReMind outperforms GRPO at all tested review budgets.}
All configurations outperform GRPO (62.83--64.06 vs.\ 60.25; Table~\ref{tab:ablation}).
Even 1\% review avoids GRPO's mid-training plateau (Figure~\ref{fig:ablation_curve}).
Increasing the budget to 4\% or 10\% yields no further gain, possibly because more review displaces fresh training signal.
At a matched 2\% budget, burst scheduling ($r\!=\!0.10$, $f\!=\!5$) outperforms smooth ($r\!=\!0.02$, $f\!=\!1$) by 0.36 points.
Accumulating reviews across several steps may yield a more diverse cohort for identifying regressed samples.
\end{minipage}
\end{figure}

\begin{table}[H]
\centering
\small
\setlength{\tabcolsep}{3.0pt}
\caption{Full results for review budget and schedule sensitivity. The upper block compares budget settings; the lower block compares burst and smooth schedules at matched $\sim$2\% budget. Shaded rows are the default; benchmark abbreviations follow Table~\ref{tab:main}.}
\label{tab:ablation}
\label{tab:ablation_full}
\resizebox{\textwidth}{!}{%
\begin{tabular}{lcc cccc c cccc c c}
\toprule
\multirow{2}{*}{\textbf{Setting}}
& \multirow{2}{*}{$r$} & \multirow{2}{*}{$f$}
& \multicolumn{5}{c}{\textbf{Math Reasoning}}
& \multicolumn{5}{c}{\textbf{General Multimodal}}
& \multirow{2}{*}{\textbf{Avg}} \\
\cmidrule(lr){4-8} \cmidrule(lr){9-13}
& & & Vista & Vision & WeMath & Verse & Avg
& MMMU-P & MMB & MM-S & ZeroB & Avg & \\
\midrule
\multicolumn{14}{l}{\textit{Budget ablation}} \\
\midrule
$\sim$1\% & 0.05 & 5
& 78.45 & 55.20 & 63.33 & 60.15 & 64.28
& 57.76 & 90.68 & 73.10 & 25.15 & 61.67 & 62.98 \\
\rowcolor{cyan!10}
$\sim$2\% (default) & 0.10 & 5
& 79.10 & 56.81 & 66.38 & 61.52 & 65.95
& 58.78 & 90.41 & 73.40 & 26.05 & 62.16 & 64.06 \\
$\sim$4\% & 0.20 & 5
& 76.50 & 55.46 & 66.00 & 60.05 & 64.50
& 57.58 & 90.84 & 73.20 & 25.90 & 61.88 & 63.19 \\
$\sim$10\% & 0.10 & 1
& 76.25 & 55.46 & 65.24 & 60.37 & 64.33
& 57.43 & 90.78 & 72.27 & 24.85 & 61.33 & 62.83 \\
\midrule
\multicolumn{14}{l}{\textit{Schedule ablation (matched $\sim$2\% budget)}} \\
\midrule
\rowcolor{cyan!10}
Burst & 0.10 & 5
& 79.10 & 56.81 & 66.38 & 61.52 & 65.95
& 58.78 & 90.41 & 73.40 & 26.05 & 62.16 & 64.06 \\
Smooth & 0.02 & 1
& 78.45 & 57.27 & 64.86 & 61.17 & 65.44
& 57.58 & 90.75 & 74.03 & 25.45 & 61.95 & 63.70 \\
\bottomrule
\end{tabular}}
\end{table}
\endgroup

\noindent\textbf{Regression-aware review outperforms random prompt review.}
We compare \method{} with two simple prompt-review strategies under the same approximately 2\% review budget.
\emph{Random-seen} samples uniformly from all previously seen prompts, while \emph{random-mastered} samples from previously mastered prompts without regression-aware queue updates.
All strategies generate fresh rollouts under the current policy.

Random-seen and random-mastered reach 62.92 and 62.40 on average, respectively, versus GRPO's 60.25, supporting the value of prompt review across selection strategies.
\method{} reaches 64.06, outperforming random-seen and random-mastered by 1.14 and 1.66 points, respectively.
Table~\ref{tab:review_strategies_full} in Appendix~\ref{app:review_strategies} reports the full eight-benchmark comparison.

\noindent\textbf{Sensitivity to the mastery criterion.}
To assess sensitivity to how mastered samples are identified, we vary the mastery criterion and report full eight-benchmark results in Appendix~\ref{app:mastery_sensitivity}.
The $K=16$ all-correct criterion reaches 64.66 on average, compared with 64.06 for $K=8$ and 60.25 for GRPO.
At $K=8$, the looser six-of-eight threshold scores 63.57, still above GRPO but below the all-correct criterion.
This fixed-$K$ comparison supports conservative mastery labeling, while gains over GRPO persist across the tested criteria.

\FloatBarrier
\subsection{Training Dynamics}
\label{sec:exp:dynamics}

\begingroup
\setlength{\intextsep}{4pt}
\begin{figure}[H]
  \centering
  \setlength{\abovecaptionskip}{3pt}
  \includegraphics[width=\columnwidth]{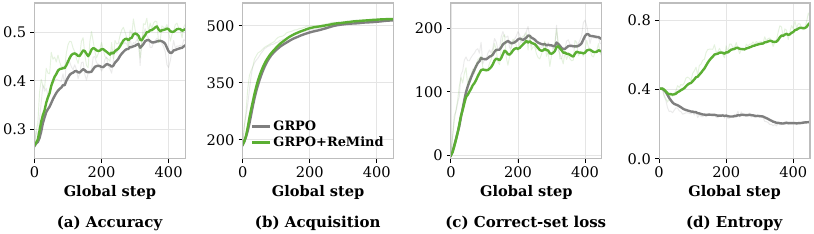}
  \caption{Training dynamics over 450 steps: (a) validation accuracy, (b) ever-solved prompts, (c) correct-set loss (count of previously solved prompts currently answered incorrectly), and (d) policy entropy. Faint lines show recorded values; solid lines show exponential moving averages.}
  \label{fig:dynamics}
\end{figure}
\endgroup

\noindent\textbf{Higher accuracy accompanies lower forgetting.}
Figure~\ref{fig:dynamics}(a--c) shows similar acquisition coverage for both methods ($\sim$500 ever-solved samples), while \method{} maintains higher accuracy and reduces correct-set loss from roughly 190 to 155 samples.
First-solve and first-mastery timing, together with matched fresh-prompt exposure, further support that review preserves acquisition (Appendix~\ref{app:acquisition_timing}).

\noindent\textbf{\method{} maintains higher policy entropy.}
In panel~(d), GRPO's entropy declines as forgetting increases between steps 50 and 150, while \method{} recovers from the initial drop and maintains entropy near 0.7--0.8.
Retaining more previously mastered solutions may help sustain a broader output distribution, with higher entropy as a by-product.

\FloatBarrier
\subsection{Further Analysis}
\label{sec:exp:further}
\label{sec:exp:efficiency}
\label{sec:exp:analysis}

\begingroup
\setlength{\intextsep}{0pt}
\begin{figure}[H]
\centering
\includegraphics[width=0.90\textwidth]{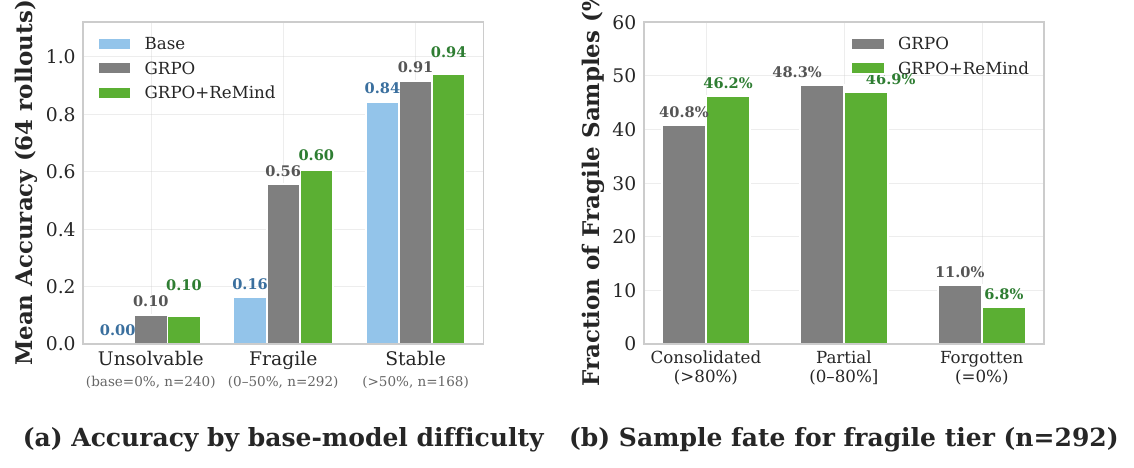}
\setlength{\abovecaptionskip}{3pt}
\caption{Retention by difficulty (64 rollouts per prompt); full breakdown in \mbox{Appendix~\ref{app:per_difficulty}}.}
\label{fig:per_difficulty}
\end{figure}
\endgroup

\noindent\textbf{Retention gains concentrate on fragile samples that are solvable but unstable.}
Figure~\ref{fig:per_difficulty} stratifies 700 validation samples into three tiers by base-model accuracy: \emph{unsolvable} (0\%, $n$=240), \emph{fragile} (0--50\%, $n$=292), and \emph{stable} ($>$50\%, $n$=168).
\method{}'s improvement over GRPO is largest on the fragile tier (+4.9\,pp) and negligible on unsolvable samples, consistent with the premise that review can only retain previously acquired knowledge.
Within the fragile tier, \method{} consolidates 46.2\% of samples to above 80\% accuracy (vs.\ 40.8\% under GRPO) and reduces complete forgetting from 11.0\% to 6.8\%, shifting fragile knowledge toward stable mastery rather than catastrophic loss.

\begingroup
\setlength{\intextsep}{2pt}
\begin{figure}[H]
\begin{minipage}[t]{0.35\textwidth}
\vspace{0pt}
\centering
\includegraphics[width=\linewidth]{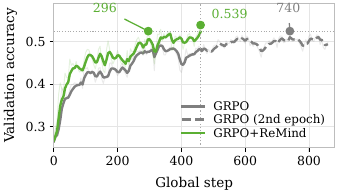}\par
\setlength{\abovecaptionskip}{3pt}
\caption{One-epoch \method{} versus two-epoch GRPO.}
\label{fig:second_epoch}
\end{minipage}\hfill
\begin{minipage}[t]{0.62\textwidth}
\vspace{0pt}
\textbf{One epoch of \method{} outperforms two epochs of GRPO.}
A natural alternative to \method{}'s timely review is simply training longer.
Figure~\ref{fig:second_epoch} compares one-epoch \method{} with two-epoch GRPO on the same data.
Second-epoch GRPO fluctuates around 0.50 and finishes at 0.497, below \method{}'s 0.539.
\method{} also matches the two-epoch peak at step 296 versus 740, using about $2.5\times$ fewer training steps.
This comparison over full training runs complements the controlled early-versus-delayed review study in \S\ref{sec:turnover:repair}, supporting the value of timely, selective review over deferring re-exposure to another epoch.
\end{minipage}
\end{figure}
\endgroup

\begingroup
\setlength{\intextsep}{2pt}
\begin{table}[H]
\begin{minipage}[t]{0.62\textwidth}
\vspace{0pt}
\noindent\textbf{The review mechanism adds negligible overhead.}
Table~\ref{tab:resource} reports wall-clock time for training, validation, and review; review accounts for only 3\% of the total.
Only a lightweight FIFO prompt queue is needed, with no additional model replicas, optimizer states, or stored historical trajectories, making \method{} easy to integrate into RLVR training loops.
\end{minipage}\hfill
\begin{minipage}[t]{0.35\textwidth}
\vspace{0pt}
\centering
\setlength{\abovecaptionskip}{0pt}
\setlength{\belowcaptionskip}{1pt}
\caption{Wall-clock breakdown.}
\label{tab:resource}
\small
\setlength{\tabcolsep}{3pt}
\renewcommand{\arraystretch}{0.9}
\begin{tabularx}{\linewidth}{@{}Xrr@{}}
\toprule
\textbf{Component} & \textbf{Time} & \textbf{Share} \\
\midrule
Training & 9.9\,h & 78\% \\
Validation & 2.5\,h & 19\% \\
\rowcolor{cyan!10}
Review & 0.3\,h & \textbf{3\%} \\
\midrule
Total & 12.7\,h & 100\% \\
\bottomrule
\end{tabularx}\par
\end{minipage}
\end{table}
\endgroup

\noindent\textbf{About 21\% of prompt reviews produce mixed outcomes.}
We logged reviews of previously mastered samples to identify learning signals.
In the lifecycle audit, 269 of 1,275 prompt-review records (21\%) contain mixed outcomes, yielding nonzero group-relative advantages.
These samples return to the queue; all-correct samples leave it.
Figure~\ref{fig:queue_dynamics} shows improving retention while some samples continue to regress.
Peak queue occupancy is 8\%, with no evictions (Appendix~\ref{app:queue_lifecycle}).

\begingroup
\setlength{\intextsep}{2pt}
\begin{figure}[H]
\centering
\includegraphics[width=\linewidth]{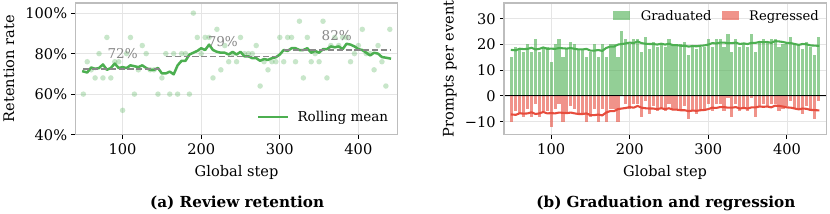}\par
\setlength{\abovecaptionskip}{3pt}
\caption{79 review events: (a) retention rates and phase means; (b) graduated and regressed counts (regressions below zero). Curves show centered nine-event means.}
\label{fig:queue_dynamics}
\end{figure}
\endgroup

\noindent\textbf{Most graduates pass later 16/16 audits, but failures remain.}
For simplicity, \method{} stops tracking samples after they pass review and leave the queue.
To test whether they regress later, we evaluate these samples under \method{} and vanilla GRPO at matched later checkpoints, using 16 rollouts per prompt.
Among samples that graduated by step 200, 11.5\% no longer achieve all-correct results under \method{}, versus 15.3\% under GRPO---a 25\% relative reduction.
Samples that graduated by step 260 show a similar pattern (Appendix~\ref{app:queue_lifecycle}).
Thus, review improves later retention, but some samples still regress after leaving the queue.
We also tested uniformly returning graduated prompts to the review queue under the same $\sim$2\% total review budget; this variant does not improve the eight-benchmark average (63.78 vs.\ 64.06; Appendix~\ref{app:recheck}).

\FloatBarrier

%% file: sections/06_conclusion.tex
\section{Conclusion}

We identify persistent \emph{correct-set turnover} during accuracy plateaus in RLVR, and show that timely review can effectively counter regression. \method{} operationalizes the timely-review principle with a lightweight, self-correcting review queue; with only 3\% additional cost, it yields consistent accuracy gains across image, video, and text reasoning with different modalities, models, and algorithmic backbones. More broadly, our results suggest that a meaningful share of the headroom in RLVR is not only about learning more, but also about losing less.  Residual turnover persists even with review, suggesting that regression can remain persistent during RLVR optimization; we hope this work encourages the community to treat sample-level retention as a primary design axis in reasoning RL, alongside reward design, exploration, and curriculum that make forgetting visible and addressable.

%% file: sections/08_ethics.tex
\section*{Ethics Statement}

This work proposes a training-time review mechanism for reinforcement learning with verifiable rewards; it does not introduce new data collection, human annotation, or model deployment, and all experiments use publicly available models and benchmarks.
\method{} is designed to improve training stability and does not alter the model's safety properties beyond what the underlying RLVR training already provides.
Because the method prioritizes retaining previously mastered samples, practitioners should monitor generalization when applying it to narrow training distributions to avoid overfitting to dataset-specific patterns.

\section*{Use of AI Assistants}

We used AI assistants for manuscript editing, code development and debugging, and assistance with analysis and interpretation of experimental results. The authors are responsible for the scientific claims, experimental results, and released code.

%% file: sections/09_appendix.tex
\section{Independent Training Runs}
\label{app:multi_seed}

We repeat GRPO+\method{} training with three additional independent seeds in the main image-text setting.
Table~\ref{tab:multi_seed} reports all eight benchmark scores for each run, together with the math, general multimodal, and overall averages.
We compute the mean and sample standard deviation across Seeds~1--3 using the unrounded averages.
The original GRPO+\method{} run is listed separately and is not included in these statistics.
The three runs achieve overall averages of 64.1575, 64.1600, and 64.2100, respectively, giving $64.18\pm0.03$ across runs.
Their math average is $66.36\pm0.27$ and general multimodal average is $61.99\pm0.24$.
All three overall averages are close to the original 64.06 result and exceed the reported GRPO reference by 3.91--3.96 points.
The base-model and GRPO rows are reference results, not repeated-run estimates.

\begin{table*}[ht]
\centering
\small
\setlength{\tabcolsep}{2.4pt}
\caption{Full image-text evaluation for three additional independent GRPO+\method{} training runs. Benchmark abbreviations follow Table~\ref{tab:main}. Math and General each average four benchmarks; Overall averages all eight. The shaded rows report means and sample standard deviations over Seeds~1--3 only, computed before rounding.}
\label{tab:multi_seed}
\resizebox{\textwidth}{!}{%
\begin{tabular}{l cccc c cccc c c}
\toprule
\multirow{2}{*}{\textbf{Method / run}}
& \multicolumn{5}{c}{\textbf{Math Reasoning}}
& \multicolumn{5}{c}{\textbf{General Multimodal}}
& \multirow{2}{*}{\textbf{Overall}} \\
\cmidrule(lr){2-6} \cmidrule(lr){7-11}
& Vista & Vision & WeMath & Verse & Avg
& MMMU-P & MMB & MM-S & ZeroB & Avg & \\
\midrule
Base Model
& 73.80 & 47.37 & 54.10 & 54.61 & 57.47
& 51.75 & 89.79 & 71.83 & 19.76 & 58.28 & 57.88 \\
GRPO
& 76.20 & 48.82 & 56.57 & 59.52 & 60.28
& 55.78 & 90.29 & 72.20 & 22.60 & 60.22 & 60.25 \\
GRPO + \method{}
& 79.10 & 56.81 & 66.38 & 61.52 & 65.95
& 58.78 & 90.41 & 73.40 & 26.05 & 62.16 & 64.06 \\
\midrule
\quad Seed 1
& 78.75 & 57.48 & 66.86 & 61.08 & 66.04
& 58.54 & 90.93 & 73.87 & 25.75 & 62.27 & 64.16 \\
\quad Seed 2
& 78.75 & 57.19 & 69.24 & 60.78 & 66.49
& 58.33 & 90.51 & 73.63 & 24.85 & 61.83 & 64.16 \\
\quad Seed 3
& 78.95 & 57.48 & 68.29 & 61.45 & 66.54
& 58.48 & 90.66 & 74.57 & 23.80 & 61.88 & 64.21 \\
\midrule
\rowcolor{cyan!10}
Mean (Seeds 1--3)
& 78.82 & 57.38 & 68.13 & 61.10 & 66.36
& 58.45 & 90.70 & 74.02 & 24.80 & 61.99 & 64.18 \\
\rowcolor{cyan!10}
Std.\ (Seeds 1--3)
& 0.12 & 0.17 & 1.20 & 0.34 & 0.27
& 0.11 & 0.21 & 0.49 & 0.98 & 0.24 & 0.03 \\
\bottomrule
\end{tabular}}
\end{table*}

\section{Full Review-Strategy Results}
\label{app:review_strategies}

Table~\ref{tab:review_strategies_full} gives the eight-benchmark breakdown for \S\ref{sec:exp:review_strategies}.
The three review strategies use the same approximately 2\% review budget and generate current-policy rollouts for the selected prompts.
Random-seen samples uniformly from previously seen prompts, while random-mastered restricts selection to previously mastered prompts without regression-aware queue updates.
Both improve the overall average over GRPO, with random-seen exceeding random-mastered by 0.52 points; restricting review to previously mastered prompts alone provides no advantage in this comparison.
\method{} achieves the highest average, with its largest advantage over either random strategy on ZeroBench.

\begin{table*}[ht]
\centering
\small
\setlength{\tabcolsep}{3.2pt}
\caption{Full results for prompt-review strategies. All methods use Qwen3-VL-8B-Instruct; benchmark abbreviations follow Table~\ref{tab:main}. Best results in each column are bold; the shaded row is \method{}.}
\label{tab:review_strategies_full}
\resizebox{\textwidth}{!}{%
\begin{tabular}{lccccccccc}
\toprule
\textbf{Method} & \textbf{Vista} & \textbf{Vision} & \textbf{WeMath} & \textbf{Verse} & \textbf{MMMU-P} & \textbf{MMB} & \textbf{MM-S} & \textbf{ZeroB} & \textbf{Avg} \\
\midrule
GRPO & 76.20 & 48.82 & 56.57 & 59.52 & 55.78 & 90.29 & 72.20 & 22.60 & 60.25 \\
Random-seen & 78.65 & 55.89 & 64.57 & 58.50 & \textbf{58.94} & 90.79 & \textbf{73.57} & 22.46 & 62.92 \\
Random-mastered & 76.95 & \textbf{56.96} & 63.81 & 60.03 & 58.01 & \textbf{91.07} & 72.43 & 19.91 & 62.40 \\
\rowcolor{cyan!10}
\method{} & \textbf{79.10} & 56.81 & \textbf{66.38} & \textbf{61.52} & 58.78 & 90.41 & 73.40 & \textbf{26.05} & \textbf{64.06} \\
\bottomrule
\end{tabular}}
\end{table*}

\clearpage
\section{Additional Sensitivity Analysis}
\label{app:ablation_full}

\subsection{Review Budget and Schedule}

Table~\ref{tab:ablation_full} in \S\ref{sec:exp:ablation} reports the full per-benchmark results for the budget and schedule configurations; we expand the analysis here.

\paragraph{Budget analysis.}
Across the four budget levels, math reasoning performance varies more than general multimodal performance (Math Avg spans 64.28--65.95 vs.\ General Avg 61.33--62.16), confirming that the review mechanism primarily benefits reasoning-intensive tasks where correct-set turnover is more frequent.
The 10\% budget ($f\!=\!1$) underperforms the 4\% budget despite allocating more capacity to review: at every step, 10\% of the batch consists of previously mastered samples, reducing the diversity of fresh training signal.
This suggests a trade-off between retention and acquisition that is well balanced at the 2\% default.

\paragraph{Schedule analysis.}
The smooth schedule (reviewing 2\% every step) achieves a slightly lower General Avg (61.95 vs.\ 62.16) and Math Avg (65.44 vs.\ 65.95) compared to burst.
Both schedules improve over GRPO, with burst review achieving a 0.36-point higher overall average at the matched review budget.

\paragraph{Training curves.}
Figure~\ref{fig:ablation_curve} compares validation accuracy under the four review budgets.
All configurations separate from GRPO early in training, complementing the final benchmark results in Table~\ref{tab:ablation_full}.

\subsection{Mastery Criterion}
\label{app:mastery_sensitivity}

Table~\ref{tab:mastery_full} reports full results for the mastery-criterion comparison in \S\ref{sec:exp:ablation}.
We compare the default 8/8 all-correct criterion, a 16/16 all-correct setting, and a looser threshold of at least six correct rollouts out of eight.
At fixed $K=8$, the all-correct criterion exceeds the looser threshold by 0.49 points on average.
The $K=16$ all-correct setting achieves the highest overall average (64.66), and every criterion improves over GRPO (60.25).

\begin{table*}[ht]
\centering
\small
\setlength{\tabcolsep}{3.2pt}
\caption{Full mastery-criterion sensitivity results with Qwen3-VL-8B-Instruct. Benchmark abbreviations follow Table~\ref{tab:main}. Best results in each column are bold; the shaded row is the default criterion.}
\label{tab:mastery_full}
\resizebox{\textwidth}{!}{%
\begin{tabular}{lccccccccc}
\toprule
\textbf{Method / criterion} & \textbf{Vista} & \textbf{Vision} & \textbf{WeMath} & \textbf{Verse} & \textbf{MMMU-P} & \textbf{MMB} & \textbf{MM-S} & \textbf{ZeroB} & \textbf{Avg} \\
\midrule
GRPO & 76.20 & 48.82 & 56.57 & 59.52 & 55.78 & 90.29 & 72.20 & 22.60 & 60.25 \\
\rowcolor{cyan!10}
GRPO + \method{}, 8/8 & 79.10 & 56.81 & 66.38 & 61.52 & 58.78 & 90.41 & 73.40 & 26.05 & 64.06 \\
GRPO + \method{}, 16/16 & 79.35 & \textbf{57.88} & \textbf{68.10} & \textbf{62.40} & \textbf{58.83} & 90.60 & 73.90 & \textbf{26.20} & \textbf{64.66} \\
GRPO + \method{}, $\geq$6/8 & \textbf{79.65} & 56.07 & 64.19 & 61.28 & 57.66 & \textbf{90.83} & \textbf{74.20} & 24.70 & 63.57 \\
\bottomrule
\end{tabular}}
\end{table*}

\section{Per-Difficulty Forgetting Breakdown}
\label{app:per_difficulty}

Table~\ref{tab:per_difficulty_full} provides a fine-grained breakdown of the per-difficulty analysis summarized in \S\ref{sec:exp:analysis}.
We roll out each of the 700 validation samples 64 times under the base model (Qwen3-VL-8B-Instruct), vanilla GRPO, and GRPO+\method{}, and stratify samples into seven bins by the base model's mean accuracy.

\begin{table*}[t]
\centering
\small
\setlength{\tabcolsep}{3.8pt}
\caption{Fine-grained per-difficulty analysis. Each row corresponds to a bin of validation samples grouped by the base model's mean accuracy over 64 rollouts. \textbf{Mean Acc.} is the average accuracy within each bin; \textbf{Regressed} reports the percentage of samples whose accuracy is lower than that of the base model.}
\label{tab:per_difficulty_full}
\begin{tabular}{l r rrr rr rr}
\toprule
\multirow{2}{*}{\textbf{Base Accuracy Bin}}
& \multirow{2}{*}{\textbf{$n$}}
& \multicolumn{3}{c}{\textbf{Mean Acc.}}
& \multicolumn{2}{c}{\textbf{$\Delta$ vs.\ GRPO}}
& \multicolumn{2}{c}{\textbf{Regressed (\%)}} \\
\cmidrule(lr){3-5} \cmidrule(lr){6-7} \cmidrule(lr){8-9}
& & Base & GRPO & \method{} & $\Delta$(Acc.) & $\Delta$(Regr.) & GRPO & \method{} \\
\midrule
$= 0\%$ & 240 & 0.000 & 0.101 & 0.096 & $-$0.006 & 0.0 & 0.0 & 0.0 \\
$(0, 10\%]$ & 137 & 0.047 & 0.443 & 0.479 & +0.036 & $-$7.3 & 24.1 & 16.8 \\
$(10, 30\%]$ & 100 & 0.187 & 0.630 & 0.689 & +0.059 & $-$4.0 & 20.0 & 16.0 \\
$(30, 50\%]$ & 55 & 0.403 & 0.699 & 0.763 & +0.064 & $-$7.3 & 21.8 & 14.5 \\
$(50, 70\%]$ & 43 & 0.591 & 0.835 & 0.902 & +0.067 & $-$7.0 & 16.3 & 9.3 \\
$(70, 90\%]$ & 38 & 0.806 & 0.859 & 0.883 & +0.024 & $-$2.6 & 21.1 & 18.4 \\
$(90, 100\%]$ & 87 & 0.981 & 0.978 & 0.981 & +0.004 & +1.1 & 12.6 & 13.8 \\
\midrule
All & 700 & 0.269 & 0.486 & 0.510 & +0.024 & $-$3.0 & 13.0 & 10.0 \\
\bottomrule
\end{tabular}
\end{table*}

\noindent\textbf{Observations.}
The improvement from \method{} over GRPO is non-uniform across difficulty levels.
\method{} yields its largest gains in the $(30, 70\%]$ base-success range (+6.4--6.7\,pp over GRPO), while base-to-final degradation is most frequent in the $(0, 10\%]$ bin.
Samples unsolved by the base model in 64 rollouts ($= 0\%$) show little difference between the two methods.

Across all 700 samples, \method{} reduces the fraction of regressed samples from 13.0\% to 10.0\%, with reductions in most difficulty bins.
For example, regression falls from 24.1\% to 16.8\% in the $(0, 10\%]$ bin and from 21.8\% to 14.5\% in the $(30, 50\%]$ bin.
In the $(90, 100\%]$ bin, the regression rate increases slightly from 12.6\% to 13.8\%, while both methods maintain near-perfect mean accuracy.

\paragraph{Mastery consolidation.}
Among the 292 fragile-tier samples ($0 < \text{base accuracy} \leq 50\%$), GRPO consolidates 119 (40.8\%) to above 80\% accuracy, while \method{} consolidates 135 (46.2\%).
Conversely, GRPO completely forgets 32 samples (11.0\%), compared with 20 (6.8\%) under \method{}.
This consolidation effect explains much of \method{}'s advantage on the downstream benchmarks: by preventing fragile knowledge from decaying to zero and instead reinforcing it toward stable mastery, the review mechanism converts partial solutions into reliable performance.

\section{Queue Lifecycle and Post-Graduation Audits}
\label{app:queue_lifecycle}

\paragraph{Review-event dynamics.}
Figure~\ref{fig:queue_dynamics} in \S\ref{sec:exp:further} tracks 79 review events, containing 1,975 prompt-review records: 1,541 pass verification and 434 regress.
These event-trace statistics are reported separately from the 1,275-record lifecycle audit below.
The per-event retention rate rises from 72\% in steps 50--150 to 82\% in steps 300--440, while the mean number of regressed samples per event falls from 6.9 to 4.6.
Late-stage reviews still contain 4--5 regressed samples per event, showing that regression persists throughout the observed run.

\paragraph{Queue lifecycle.}
We log queue updates during training to examine repeated review, graduation, and capacity usage.
A reviewed prompt that regresses is re-enqueued for another review, while a prompt that passes the current-policy all-correct check graduates and leaves the queue.
Table~\ref{tab:queue_lifecycle} summarizes the lifecycle audit: 269 of 1,275 prompt-review records contain mixed correct and incorrect rollouts (21.1\%) and trigger re-enqueueing.
The denominator counts review records, including repeat reviews, rather than distinct prompts.
Of the 1,006 graduated prompts, 65 had previously regressed before passing a later review; median queue residence among graduates is 71 training steps.
These records show that the queue supports repeated review of unstable prompts rather than treating the initial mastery observation as a permanent label.
Peak occupancy was only 8\% of the 20,000-prompt capacity, and no capacity-driven evictions occurred.

\begin{table}[ht]
\centering
\small
\caption{Queue lifecycle audit for \method{}. Review counts refer to prompt-review records; graduation counts refer to distinct prompts.}
\label{tab:queue_lifecycle}
\begin{tabular}{lr}
\toprule
\textbf{Statistic} & \textbf{Value} \\
\midrule
Prompt-review records & 1,275 \\
Partial regression and re-enqueueing & 269 (21.1\%) \\
Graduated prompts & 1,006 \\
Graduated prompts with a prior regression & 65 \\
Median queue residence of graduates (steps) & 71 \\
Queue capacity (prompts) & 20,000 \\
Peak occupancy (\% of capacity) & 8\% \\
Capacity-driven evictions & 0 \\
\bottomrule
\end{tabular}
\end{table}

\paragraph{Post-graduation retention.}
We audit cohorts of prompts that graduated by training steps 200 and 260 at later checkpoints, using 16 rollouts per prompt.
Strict retention requires all 16 rollouts to be correct; strict non-retention is the fraction of audited prompts that do not meet this criterion.
For each cohort, we compare \method{} and vanilla GRPO at the same audit checkpoint.
Table~\ref{tab:post_graduation} shows that \method{} lowers strict non-retention from 15.3\% to 11.5\% for the by-step-200 cohort and from 16.4\% to 12.3\% for the by-step-260 cohort, approximately a 25\% relative reduction in both cases.
Graduation confirms an all-correct outcome at the review checkpoint; it does not guarantee permanent retention.
The remaining 11.5--12.3\% strict non-retention shows that later regression persists among graduated prompts even as \method{} improves their retention relative to GRPO.

\begin{table}[ht]
\centering
\small
\caption{Post-graduation audits with 16 rollouts per prompt. Entries are strict non-retention rates (\%, lower is better); the two methods are evaluated at the same later checkpoint for each cohort.}
\label{tab:post_graduation}
\begin{tabular}{lrr}
\toprule
\textbf{Graduated cohort} & \textbf{GRPO} & \textbf{GRPO + \method{}} \\
\midrule
By step 200 & 15.3 & \textbf{11.5} \\
By step 260 & 16.4 & \textbf{12.3} \\
\bottomrule
\end{tabular}
\end{table}

\FloatBarrier
\subsection{Additional Review of Graduated Prompts}
\label{app:recheck}

We evaluate a variant that periodically re-enqueues graduated prompts for review and training, using uniform rather than risk-based selection.
The total review budget remains approximately 2\%, so these additional reviews share the budget allocated to the review queue.
Table~\ref{tab:recheck_full} reports the full eight-benchmark evaluation with Qwen3-VL-8B-Instruct.
This variant reaches 63.78 overall, compared with 64.06 for standard \method{} and 60.25 for GRPO.
It improves some benchmark scores but provides no overall gain over \method{} in this comparison.

\begin{table*}[ht]
\centering
\small
\setlength{\tabcolsep}{2.4pt}
\caption{Additional review of graduated prompts at the same approximately 2\% total review budget as standard \method{}. Benchmark abbreviations follow Table~\ref{tab:main}. Math and General average their four respective benchmarks; Overall averages all eight.}
\label{tab:recheck_full}
\resizebox{\textwidth}{!}{%
\begin{tabular}{l cccc c cccc c c}
\toprule
\multirow{2}{*}{\textbf{Method}}
& \multicolumn{5}{c}{\textbf{Math Reasoning}}
& \multicolumn{5}{c}{\textbf{General Multimodal}}
& \multirow{2}{*}{\textbf{Overall}} \\
\cmidrule(lr){2-6} \cmidrule(lr){7-11}
& Vista & Vision & WeMath & Verse & Avg
& MMMU-P & MMB & MM-S & ZeroB & Avg & \\
\midrule
Base Model
& 73.80 & 47.37 & 54.10 & 54.61 & 57.47
& 51.75 & 89.79 & 71.83 & 19.76 & 58.28 & 57.88 \\
GRPO
& 76.20 & 48.82 & 56.57 & 59.52 & 60.28
& 55.78 & 90.29 & 72.20 & 22.60 & 60.22 & 60.25 \\
\rowcolor{cyan!10}
GRPO + \method{}
& 79.10 & 56.81 & 66.38 & 61.52 & 65.95
& 58.78 & 90.41 & 73.40 & 26.05 & 62.16 & 64.06 \\
\quad w/ additional review
& 77.95 & 56.68 & 67.05 & 61.37 & 65.76
& 58.52 & 90.61 & 74.10 & 23.95 & 61.80 & 63.78 \\
\bottomrule
\end{tabular}}
\end{table*}

A possible explanation is that uniform additional review of graduated prompts diverts budget toward prompts that remain stable.
In the by-step-200 audit, 88.5\% of graduated prompts retain all-correct performance across 16 rollouts; allocating additional reviews to these prompts can reduce opportunities to revisit regressed prompts already in the queue.
This interpretation is consistent with the fixed-budget result, but the aggregate scores do not directly measure budget displacement or changes in post-graduation retention.
An adaptive schedule that prioritizes graduates at higher risk of regression is a direction for future work.

\FloatBarrier
\clearpage
\section{Acquisition Timing and Fresh-Prompt Exposure}
\label{app:acquisition_timing}

\paragraph{First-solve and first-mastery timing.}
To complement the cumulative ever-solved counts in \S\ref{sec:exp:dynamics}, we examine when prompts are first solved and first meet the all-correct mastery criterion.
Table~\ref{tab:acquisition_timing} reports the median and 90th percentile of the recorded first-solve and first-mastery steps.
With \method{}, median first-solve time decreases from 163 to 158 steps and median first-mastery time from 192 to 189 steps.
The corresponding 90th percentiles also decrease, from 278 to 272 and from 286 to 280 steps.
Thus, neither reported timing quantile indicates delayed acquisition under review.

\begin{table}[H]
\centering
\small
\caption{Per-prompt acquisition timing in training steps (lower is earlier). The 90th percentile is denoted by p90.}
\label{tab:acquisition_timing}
\begin{tabular}{lrrrr}
\toprule
\multirow{2}{*}{\textbf{Metric}} & \multicolumn{2}{c}{\textbf{Median}} & \multicolumn{2}{c}{\textbf{p90}} \\
\cmidrule(lr){2-3} \cmidrule(lr){4-5}
& GRPO & \method{} & GRPO & \method{} \\
\midrule
First solve & 163 & 158 & 278 & 272 \\
First mastery & 192 & 189 & 286 & 280 \\
\bottomrule
\end{tabular}
\end{table}

\paragraph{Matched fresh-prompt exposure.}
Because review replaces approximately 2\% of training prompts, we also compare acquisition after both methods have seen the same number of fresh prompts.
At this matched exposure, \method{} has solved 242 more prompts and mastered 368 more prompts than GRPO.
This comparison complements the timing statistics by aligning the amount of fresh-prompt exposure, and both analyses support that the default review budget preserves acquisition in this setting.

\section{Cross-Modal Evaluation Benchmarks}
\label{app:video_benchmarks}

Table~\ref{tab:video_benchmarks} summarizes the six video understanding benchmarks used in \S\ref{sec:exp:main}.
They cover general perception, long-video comprehension, video reasoning, and STEM knowledge.
For text math reasoning, we evaluate on AIME 2024/2025, AMC~\citep{li2024numinamath}, MATH-500~\citep{hendrycks2021measuringmathematicalproblemsolving}, Minerva~\citep{dataset_minerva}, and OlympiadBench~\citep{dataset_olympiad}.

\begin{table}[h]
\centering
\small
\caption{Video understanding benchmarks used in the cross-modal evaluation (\S\ref{sec:exp:main}).}
\label{tab:video_benchmarks}
\begin{tabular}{lrr}
\toprule
\textbf{Benchmark} & \textbf{Samples} & \textbf{Metric} \\
\midrule
\multicolumn{3}{l}{\textit{General Video Understanding}} \\
MVBench~\citep{li2024mvbenchcomprehensivemultimodalvideo} & 3{,}586 & Accuracy \\
MotionBench~\citep{hong2025motionbenchbenchmarkingimprovingfinegrained} & 3{,}715 & Accuracy \\
\midrule
\multicolumn{3}{l}{\textit{Long Video Understanding}} \\
MLVU~\citep{zhou2025mlvubenchmarkingmultitasklong} & 502 & Accuracy \\
\midrule
\multicolumn{3}{l}{\textit{Video Reasoning}} \\
Video-Holmes~\citep{cheng2025videoholmesmllmthinklike} & 1{,}837 & Accuracy \\
LongVideo-Reason~\citep{chen2025scalingrllongvideos} & 851 & Accuracy \\
\midrule
\multicolumn{3}{l}{\textit{STEM Knowledge}} \\
VideoMathQA~\citep{rasheed2025videomathqabenchmarkingmathematicalreasoning} & 2{,}100 & Accuracy \\
\bottomrule
\end{tabular}
\end{table}

\clearpage
\setlength{\parskip}{3pt}
\section{Experimental Setup Details}
\label{app:setup_details}

\paragraph{Training data.}
MMFineReason-123K~\citep{lin2026mmfinereasonclosingmultimodalreasoning} is derived from the MMFineReason-1.8M corpus via difficulty-based filtering.
Each candidate sample is rolled out four times with Qwen3-VL-4B-Thinking~\citep{qwen3vl}, and only samples on which the model fails every attempt are retained, yielding a training set of 123K challenging multimodal reasoning problems.

\paragraph{Evaluation benchmarks.}
We evaluate on eight multimodal reasoning benchmarks spanning visually grounded mathematics, broad subject knowledge, and fine-grained visual understanding.
MathVista~\citep{lumathvista}, MathVision~\citep{wang2024measuring}, WeMath~\citep{qiao2025we}, and MathVerse~\citep{zhang2024mathverse} target mathematical reasoning in visual contexts, from general problem solving to competition-level diagrams and structured difficulty tiers.
MMMU-Pro~\citep{yue2025mmmu} and MMBench~\citep{liu2024mmbench} test broad multi-discipline knowledge and general multimodal competence.
MM-Star~\citep{chen2024we} stresses fine-grained visual discrimination, and ZeroBench~\citep{roberts2025zerobench} serves as a hard stress test for the tail of the reasoning distribution.

\paragraph{Baselines.}
All baselines use the same base policy and GRPO-style training framework, with method-specific objectives and sampling rules.
GRPO~\citep{shao2024deepseekmath} is the pure on-policy reference.
DAPO~\citep{yu2025dapo} augments GRPO with dynamic sampling and online filtering.
MCPO~\citep{liao2026mcpo} regularizes policy drift on mastered prompts through hinge-KL regularization and reweights majority-correct prompts.
RL-ZVP~\citep{le2026rlzvp} constructs entropy-guided advantages for zero-variance groups, retaining the GRPO formulation for groups with mixed outcomes.
VCRL~\citep{jiang2025vcrl} uses a variance-based curriculum for prompt selection.
We combine \method{} with both DAPO and VCRL to evaluate its compatibility with dynamic sampling and curriculum learning, using the default approximately 2\% review budget.
ExGRPO~\citep{liang2025exgrpo} reuses successful trajectories from historical replay buffers.
RLEP~\citep{zhang2025rlep} replays trajectories from a far-future checkpoint.
RePO~\citep{li2025repo} collects early on-policy rollouts and revisits them asynchronously.
We also report the base model (Qwen3-VL-8B-Instruct) without RL post-training as a reference point.

\paragraph{Implementation details.}
Our implementation is built on the EasyVideoR1~\citep{qin2026easyvideor1easierrlvideo} framework.
The maximum context length is 8192 tokens, split into a prompt budget of 4096 and a response budget of 4096, applied consistently at training and evaluation time.
We train with a learning rate of $1 \times 10^{-6}$ and a batch size of 256, sampling $K\!=\!8$ rollouts per prompt at temperature $1.0$.
Clipping thresholds are set to $\epsilon_{\text{low}}=0.2$ and $\epsilon_{\text{high}}=0.28$; the GRPO backbone omits the standard KL penalty and entropy regularization.
For \method{}, we set the review ratio $r\!=\!0.10$, review period $f\!=\!5$, and enqueue probability $p_{\mathrm{add}}\!=\!0.25$.
Under these settings, review samples constitute approximately 2\% of total training prompts on average.
All experiments are run on 4 compute nodes, each with 8 NVIDIA H200 140GB GPUs.

\section{Training-Time Breakdown}
\label{app:runtime}

Table~\ref{tab:resource} in \S\ref{sec:exp:efficiency} reports timings for the 12.7\,h training run.
Standard training includes rollout generation and policy updates, and validation is performed periodically.
Review overhead is measured as the marginal step-time increase on review steps relative to non-review steps, covering review rollouts and queue management.
Review requires only a prompt queue, without additional model replicas, optimizer states, or stored historical trajectories.

%% file: sections/07_limitations.tex
\section{Limitations}
\label{app:limitations}

\method{} inherits the standard RLVR assumption that the verifiable reward is correct; because mastery status directly governs which samples enter the review queue, systematic reward errors (e.g., false positives from imprecise answer extraction) would be more consequential here than in methods that do not condition on mastery.
Our experiments cover 7--8B-scale models on mathematical and multimodal reasoning; whether the optimal review budget or the magnitude of retention gains transfers to substantially larger models or other domains such as code generation remains to be verified.
\method{} reduces but does not eliminate correct-set regression; our current design uses a fixed review schedule, and adaptive scheduling that responds to the model's regression dynamics could further narrow this gap.

%% file: main.bbl
\begin{thebibliography}{49}
\providecommand{\natexlab}[1]{#1}
\providecommand{\url}[1]{\texttt{#1}}
\expandafter\ifx\csname urlstyle\endcsname\relax
  \providecommand{\doi}[1]{doi: #1}\else
  \providecommand{\doi}{doi: \begingroup \urlstyle{rm}\Url}\fi

\bibitem[Bae et~al.(2025)Bae, Hong, Lee, Kim, Nam, and Kwak]{bae2025online}
Sanghwan Bae, Jiwoo Hong, Min~Young Lee, Hanbyul Kim, JeongYeon Nam, and
  Donghyun Kwak.
\newblock Online difficulty filtering for reasoning oriented reinforcement
  learning.
\newblock \emph{arXiv preprint arXiv:2504.03380}, 2025.

\bibitem[Bai et~al.(2025)Bai, Cai, Chen, Chen, Chen, Cheng, Deng, Ding, Gao,
  Ge, Ge, Guo, Huang, Huang, Huang, Hui, Jiang, Li, Li, Li, Li, Lin, Lin, Liu,
  Liu, Liu, Liu, Liu, Liu, Lu, Luo, Lv, Men, Meng, Ren, Ren, Song, Sun, Tang,
  Tu, Wan, Wang, Wang, Wang, Wang, Xie, Xu, Xu, Xu, Yang, Yang, Yang, Yang, Yu,
  Zhang, Zhang, Zhang, Zheng, Zhong, Zhou, Zhou, Zhou, Zhu, and Zhu]{qwen3vl}
Shuai Bai, Yuxuan Cai, Ruizhe Chen, Keqin Chen, Xionghui Chen, Zesen Cheng,
  Lianghao Deng, Wei Ding, Chang Gao, Chunjiang Ge, Wenbin Ge, Zhifang Guo,
  Qidong Huang, Jie Huang, Fei Huang, Binyuan Hui, Shutong Jiang, Zhaohai Li,
  Mingsheng Li, Mei Li, Kaixin Li, Zicheng Lin, Junyang Lin, Xuejing Liu,
  Jiawei Liu, Chenglong Liu, Yang Liu, Dayiheng Liu, Shixuan Liu, Dunjie Lu,
  Ruilin Luo, Chenxu Lv, Rui Men, Lingchen Meng, Xuancheng Ren, Xingzhang Ren,
  Sibo Song, Yuchong Sun, Jun Tang, Jianhong Tu, Jianqiang Wan, Peng Wang,
  Pengfei Wang, Qiuyue Wang, Yuxuan Wang, Tianbao Xie, Yiheng Xu, Haiyang Xu,
  Jin Xu, Zhibo Yang, Mingkun Yang, Jianxin Yang, An~Yang, Bowen Yu, Fei Zhang,
  Hang Zhang, Xi~Zhang, Bo~Zheng, Humen Zhong, Jingren Zhou, Fan Zhou, Jing
  Zhou, Yuanzhi Zhu, and Ke~Zhu.
\newblock Qwen3-vl technical report, 2025.
\newblock URL \url{https://arxiv.org/abs/2511.21631}.

\bibitem[Cepeda et~al.(2006)Cepeda, Pashler, Vul, Wixted, and
  Rohrer]{cepeda2006distributed}
Nicholas~J Cepeda, Harold Pashler, Edward Vul, John~T Wixted, and Doug Rohrer.
\newblock Distributed practice in verbal recall tasks: A review and
  quantitative synthesis.
\newblock \emph{Psychological bulletin}, 132\penalty0 (3):\penalty0 354, 2006.

\bibitem[Chen et~al.(2024)Chen, Li, Dong, Zhang, Zang, Chen, Duan, Wang, Qiao,
  Lin, et~al.]{chen2024we}
Lin Chen, Jinsong Li, Xiaoyi Dong, Pan Zhang, Yuhang Zang, Zehui Chen, Haodong
  Duan, Jiaqi Wang, Yu~Qiao, Dahua Lin, et~al.
\newblock Are we on the right way for evaluating large vision-language models?
\newblock \emph{Advances in Neural Information Processing Systems},
  37:\penalty0 27056--27087, 2024.

\bibitem[Chen et~al.(2025)Chen, Huang, Shi, Hu, Ye, Zhu, Liu, Molchanov, Kautz,
  Qi, Liu, Yin, Lu, and Han]{chen2025scalingrllongvideos}
Yukang Chen, Wei Huang, Baifeng Shi, Qinghao Hu, Hanrong Ye, Ligeng Zhu,
  Zhijian Liu, Pavlo Molchanov, Jan Kautz, Xiaojuan Qi, Sifei Liu, Hongxu Yin,
  Yao Lu, and Song Han.
\newblock Scaling rl to long videos, 2025.
\newblock URL \url{https://arxiv.org/abs/2507.07966}.

\bibitem[Cheng et~al.(2025)Cheng, Ge, Wang, Ge, Liao, and
  Shan]{cheng2025videoholmesmllmthinklike}
Junhao Cheng, Yuying Ge, Teng Wang, Yixiao Ge, Jing Liao, and Ying Shan.
\newblock Video-holmes: Can mllm think like holmes for complex video
  reasoning?, 2025.
\newblock URL \url{https://arxiv.org/abs/2505.21374}.

\bibitem[Cui et~al.(2025)Cui, Zhang, Chen, Yuan, Wang, Zuo, Li, Fan, Chen,
  Chen, et~al.]{cui2025entropy}
Ganqu Cui, Yuchen Zhang, Jiacheng Chen, Lifan Yuan, Zhi Wang, Yuxin Zuo,
  Haozhan Li, Yuchen Fan, Huayu Chen, Weize Chen, et~al.
\newblock The entropy mechanism of reinforcement learning for reasoning
  language models.
\newblock \emph{arXiv preprint arXiv:2505.22617}, 2025.

\bibitem[Dong et~al.(2026)Dong, Fu, Li, Zhu, Liu, Shao, Ye, and
  Wang]{dong2026probing}
Yiming Dong, Kun Fu, Haoyu Li, Xinyuan Zhu, Yurou Liu, Lijing Shao, Jieping Ye,
  and Zheng Wang.
\newblock Probing {RLVR} training instability through the lens of
  objective-level hacking.
\newblock \emph{arXiv preprint arXiv:2602.01103}, 2026.

\bibitem[Dou et~al.(2025)Dou, Wu, Xu, Zheng, Gui, Zhang, and Huang]{dou2025rrl}
Shihan Dou, Muling Wu, Jingwen Xu, Rui Zheng, Tao Gui, Qi~Zhang, and Xuanjing
  Huang.
\newblock Improving {RL} exploration for {LLM} reasoning through retrospective
  replay.
\newblock \emph{arXiv preprint arXiv:2504.14363}, 2025.

\bibitem[Ebbinghaus(1885)]{ebbinghaus1885memory}
Hermann Ebbinghaus.
\newblock \emph{{\"U}ber das ged{\"a}chtnis: untersuchungen zur experimentellen
  psychologie}.
\newblock Duncker \& Humblot, 1885.

\bibitem[Guo et~al.(2025)Guo, Yang, Zhang, Song, Zhang, Xu, Zhu, Ma, Wang, Bi,
  et~al.]{guo2025deepseek}
Daya Guo, Dejian Yang, Haowei Zhang, Junxiao Song, Ruoyu Zhang, Runxin Xu,
  Qihao Zhu, Shirong Ma, Peiyi Wang, Xiao Bi, et~al.
\newblock Deepseek-{R}1: Incentivizing reasoning capability in llms via
  reinforcement learning.
\newblock \emph{arXiv preprint arXiv:2501.12948}, 2025.

\bibitem[He et~al.(2024)He, Luo, Bai, Hu, Thai, Shen, Hu, Han, Huang, Zhang,
  et~al.]{dataset_olympiad}
Chaoqun He, Renjie Luo, Yuzhuo Bai, Shengding Hu, Zhen Thai, Junhao Shen, Jinyi
  Hu, Xu~Han, Yujie Huang, Yuxiang Zhang, et~al.
\newblock Olympiadbench: A challenging benchmark for promoting agi with
  olympiad-level bilingual multimodal scientific problems.
\newblock In \emph{Proceedings of the 62nd Annual Meeting of the Association
  for Computational Linguistics (Volume 1: Long Papers)}, pp.\  3828--3850,
  2024.

\bibitem[Hendrycks et~al.(2021)Hendrycks, Burns, Kadavath, Arora, Basart, Tang,
  Song, and Steinhardt]{hendrycks2021measuringmathematicalproblemsolving}
Dan Hendrycks, Collin Burns, Saurav Kadavath, Akul Arora, Steven Basart, Eric
  Tang, Dawn Song, and Jacob Steinhardt.
\newblock Measuring mathematical problem solving with the math dataset, 2021.
\newblock URL \url{https://arxiv.org/abs/2103.03874}.

\bibitem[Hong et~al.(2025)Hong, Cheng, Yang, Wang, Wang, Gu, Huang, Dong, and
  Tang]{hong2025motionbenchbenchmarkingimprovingfinegrained}
Wenyi Hong, Yean Cheng, Zhuoyi Yang, Weihan Wang, Lefan Wang, Xiaotao Gu, Shiyu
  Huang, Yuxiao Dong, and Jie Tang.
\newblock Motionbench: Benchmarking and improving fine-grained video motion
  understanding for vision language models, 2025.
\newblock URL \url{https://arxiv.org/abs/2501.02955}.

\bibitem[Hu et~al.(2026)Hu, Cai, Huang, Yao, Zhang, Zhang, Deng, and
  Chen]{hu2026emergentslowthinkingllms}
Sihan Hu, Xiansheng Cai, Yuan Huang, Zhiyuan Yao, Linfeng Zhang, Pan Zhang,
  Youjin Deng, and Kun Chen.
\newblock Emergent slow thinking in llms as inverse tree freezing, 2026.
\newblock URL \url{https://arxiv.org/abs/2509.23629}.

\bibitem[Jiang et~al.(2025)Jiang, Feng, Quan, Hao, Zhang, Liu, and
  Wang]{jiang2025vcrl}
Guochao Jiang, Wenfeng Feng, Guofeng Quan, Chuzhan Hao, Yuewei Zhang, Guohua
  Liu, and Hao Wang.
\newblock {VCRL}: Variance-based curriculum reinforcement learning for large
  language models.
\newblock \emph{arXiv preprint arXiv:2509.19803}, 2025.

\bibitem[Kirkpatrick et~al.(2017)Kirkpatrick, Pascanu, Rabinowitz, Veness,
  Desjardins, Rusu, Milan, Quan, Ramalho, Grabska-Barwinska, Hassabis, Clopath,
  Kumaran, and Hadsell]{kirkpatrick2017overcoming}
James Kirkpatrick, Razvan Pascanu, Neil Rabinowitz, Joel Veness, Guillaume
  Desjardins, Andrei~A. Rusu, Kieran Milan, John Quan, Tiago Ramalho, Agnieszka
  Grabska-Barwinska, Demis Hassabis, Claudia Clopath, Dharshan Kumaran, and
  Raia Hadsell.
\newblock Overcoming catastrophic forgetting in neural networks.
\newblock \emph{Proceedings of the National Academy of Sciences}, 114\penalty0
  (13):\penalty0 3521–3526, March 2017.
\newblock ISSN 1091-6490.
\newblock \doi{10.1073/pnas.1611835114}.
\newblock URL \url{http://dx.doi.org/10.1073/pnas.1611835114}.

\bibitem[Le et~al.(2026)Le, Jeon, Vu, Lai, and Yang]{le2026rlzvp}
Thanh-Long~V. Le, Myeongho Jeon, Kim Vu, Viet Lai, and Eunho Yang.
\newblock No prompt left behind: Exploiting zero-variance prompts in {LLM}
  reinforcement learning via entropy-guided advantage shaping.
\newblock In \emph{The Fourteenth International Conference on Learning
  Representations}, 2026.
\newblock URL \url{https://openreview.net/forum?id=kiXFIESZKv}.

\bibitem[Lewkowycz et~al.(2022)Lewkowycz, Andreassen, Dohan, Dyer, Michalewski,
  Ramasesh, Slone, Anil, Schlag, Gutman-Solo, et~al.]{dataset_minerva}
Aitor Lewkowycz, Anders Andreassen, David Dohan, Ethan Dyer, Henryk
  Michalewski, Vinay Ramasesh, Ambrose Slone, Cem Anil, Imanol Schlag, Theo
  Gutman-Solo, et~al.
\newblock Solving quantitative reasoning problems with language models.
\newblock \emph{Advances in Neural Information Processing Systems},
  35:\penalty0 3843--3857, 2022.

\bibitem[Li et~al.(2024{\natexlab{a}})Li, Beeching, Tunstall, Lipkin,
  Soletskyi, Huang, Rasul, Yu, Jiang, Shen, et~al.]{li2024numinamath}
Jia Li, Edward Beeching, Lewis Tunstall, Ben Lipkin, Roman Soletskyi, Shengyi
  Huang, Kashif Rasul, Longhui Yu, Albert~Q Jiang, Ziju Shen, et~al.
\newblock Numinamath: The largest public dataset in ai4maths with 860k pairs of
  competition math problems and solutions.
\newblock \emph{Hugging Face repository}, 2024{\natexlab{a}}.

\bibitem[Li et~al.(2024{\natexlab{b}})Li, Wang, He, Li, Wang, Liu, Wang, Xu,
  Chen, Luo, Wang, and Qiao]{li2024mvbenchcomprehensivemultimodalvideo}
Kunchang Li, Yali Wang, Yinan He, Yizhuo Li, Yi~Wang, Yi~Liu, Zun Wang, Jilan
  Xu, Guo Chen, Ping Luo, Limin Wang, and Yu~Qiao.
\newblock Mvbench: A comprehensive multi-modal video understanding benchmark,
  2024{\natexlab{b}}.
\newblock URL \url{https://arxiv.org/abs/2311.17005}.

\bibitem[Li et~al.(2025)Li, Zhou, Lam, Yang, and Lu]{li2025repo}
Siheng Li, Zhanhui Zhou, Wai Lam, Chao Yang, and Chaochao Lu.
\newblock Repo: Replay-enhanced policy optimization.
\newblock \emph{arXiv preprint arXiv:2506.09340}, 2025.

\bibitem[Liao et~al.(2026)Liao, Gao, Yang, Hu, and Ding]{liao2026mcpo}
Zhaokang Liao, Yingguo Gao, Yi~Yang, Yongheng Hu, and Jingting Ding.
\newblock {MCPO}: Mastery-consolidated policy optimization for large reasoning
  models.
\newblock \emph{arXiv preprint arXiv:2604.16972}, 2026.
\newblock URL \url{https://arxiv.org/abs/2604.16972}.

\bibitem[Lin et~al.(2026)Lin, Liu, Zhu, Qin, Lin, Shang, He, Zhang, and
  Wu]{lin2026mmfinereasonclosingmultimodalreasoning}
Honglin Lin, Zheng Liu, Yun Zhu, Chonghan Qin, Juekai Lin, Xiaoran Shang,
  Conghui He, Wentao Zhang, and Lijun Wu.
\newblock Mmfinereason: Closing the multimodal reasoning gap via open
  data-centric methods, 2026.
\newblock URL \url{https://arxiv.org/abs/2601.21821}.

\bibitem[Lin(1992)]{lin1992experience}
Long-Ji Lin.
\newblock Self-improving reactive agents based on reinforcement learning,
  planning and teaching.
\newblock \emph{Machine Learning}, 8\penalty0 (3--4):\penalty0 293--321, 1992.

\bibitem[Liu et~al.(2024)Liu, Duan, Zhang, Li, Zhang, Zhao, Yuan, Wang, He,
  Liu, et~al.]{liu2024mmbench}
Yuan Liu, Haodong Duan, Yuanhan Zhang, Bo~Li, Songyang Zhang, Wangbo Zhao, Yike
  Yuan, Jiaqi Wang, Conghui He, Ziwei Liu, et~al.
\newblock Mmbench: Is your multi-modal model an all-around player?
\newblock In \emph{European conference on computer vision}, pp.\  216--233.
  Springer, 2024.

\bibitem[Liu et~al.(2025)Liu, Chen, Li, Qi, Pang, Du, Lee, and
  Lin]{liu2025understanding}
Zichen Liu, Changyu Chen, Wenjun Li, Penghui Qi, Tianyu Pang, Chao Du, Wee~Sun
  Lee, and Min Lin.
\newblock Understanding {R1}-zero-like training: A critical perspective.
\newblock \emph{arXiv preprint arXiv:2503.20783}, 2025.

\bibitem[Lu et~al.(2024)Lu, Bansal, Xia, Liu, Li, Hajishirzi, Cheng, Chang,
  Galley, and Gao]{lumathvista}
Pan Lu, Hritik Bansal, Tony Xia, Jiacheng Liu, Chunyuan Li, Hannaneh
  Hajishirzi, Hao Cheng, Kai-Wei Chang, Michel Galley, and Jianfeng Gao.
\newblock Mathvista: Evaluating mathematical reasoning of foundation models in
  visual contexts.
\newblock In \emph{The Twelfth International Conference on Learning
  Representations}, 2024.

\bibitem[Ma et~al.(2026)Ma, Zeng, Song, Cui, Zhao, Liu, and
  Elhoseiny]{ma2026freshper}
Weiyu Ma, Yongcheng Zeng, Yan Song, Xinyu Cui, Jian Zhao, Xuhui Liu, and
  Mohamed Elhoseiny.
\newblock Freshness-aware prioritized experience replay for {LLM}/{VLM}
  reinforcement learning.
\newblock \emph{arXiv preprint arXiv:2604.16918}, 2026.

\bibitem[Mnih et~al.(2015)Mnih, Kavukcuoglu, Silver, Rusu, Veness, Bellemare,
  Graves, Riedmiller, Fidjeland, Ostrovski, et~al.]{mnih2015human}
Volodymyr Mnih, Koray Kavukcuoglu, David Silver, Andrei~A. Rusu, Joel Veness,
  Marc~G. Bellemare, Alex Graves, Martin Riedmiller, Andreas~K. Fidjeland,
  Georg Ostrovski, et~al.
\newblock Human-level control through deep reinforcement learning.
\newblock \emph{Nature}, 518\penalty0 (7540):\penalty0 529--533, 2015.

\bibitem[Phan et~al.(2025)Phan, Yang, Yao, Zhang, Bi, Tang, Khabsa, Liu, and
  Lei]{phan2025beyond}
Hoang Phan, Xianjun Yang, Kevin Yao, Jingyu Zhang, Shengjie Bi, Xiaocheng Tang,
  Madian Khabsa, Lijuan Liu, and Deren Lei.
\newblock Beyond reasoning gains: Mitigating general capabilities forgetting in
  large reasoning models.
\newblock \emph{arXiv preprint arXiv:2510.21978}, 2025.

\bibitem[Qiao et~al.(2025)Qiao, Tan, Dong, MinhuiWu, Sun, Song, Wang, Gongque,
  Lei, Zhang, et~al.]{qiao2025we}
Runqi Qiao, Qiuna Tan, Guanting Dong, MinhuiWu MinhuiWu, Chong Sun, Xiaoshuai
  Song, Jiapeng Wang, Zhuoma Gongque, Shanglin Lei, Yifan Zhang, et~al.
\newblock We-math: Does your large multimodal model achieve human-like
  mathematical reasoning?
\newblock In \emph{Proceedings of the 63rd Annual Meeting of the Association
  for Computational Linguistics (Volume 1: Long Papers)}, pp.\  20023--20070,
  2025.

\bibitem[Qin et~al.(2026)Qin, Yang, Si, Gu, Yao, Lin, Fu, Duan, and
  Wang]{qin2026easyvideor1easierrlvideo}
Chuanyu Qin, Chenxu Yang, Qingyi Si, Naibin Gu, Dingyu Yao, Zheng Lin, Peng Fu,
  Nan Duan, and Jiaqi Wang.
\newblock Easyvideor1: Easier rl for video understanding, 2026.
\newblock URL \url{https://arxiv.org/abs/2604.16893}.

\bibitem[Rasheed et~al.(2025)Rasheed, Shaker, Tang, Maaz, Yang, Khan, and
  Khan]{rasheed2025videomathqabenchmarkingmathematicalreasoning}
Hanoona Rasheed, Abdelrahman Shaker, Anqi Tang, Muhammad Maaz, Ming-Hsuan Yang,
  Salman Khan, and Fahad~Shahbaz Khan.
\newblock Videomathqa: Benchmarking mathematical reasoning via multimodal
  understanding in videos, 2025.
\newblock URL \url{https://arxiv.org/abs/2506.05349}.

\bibitem[Rebuffi et~al.(2017)Rebuffi, Kolesnikov, Sperl, and
  Lampert]{rebuffi2017icarl}
Sylvestre-Alvise Rebuffi, Alexander Kolesnikov, Georg Sperl, and Christoph~H.
  Lampert.
\newblock icarl: Incremental classifier and representation learning, 2017.
\newblock URL \url{https://arxiv.org/abs/1611.07725}.

\bibitem[Roberts et~al.(2025)Roberts, Taesiri, Sharma, Gupta, Roberts,
  Croitoru, Bogolin, Tang, Langer, Raina, et~al.]{roberts2025zerobench}
Jonathan Roberts, Mohammad~Reza Taesiri, Ansh Sharma, Akash Gupta, Samuel
  Roberts, Ioana Croitoru, Simion-Vlad Bogolin, Jialu Tang, Florian Langer,
  Vyas Raina, et~al.
\newblock Zerobench: An impossible visual benchmark for contemporary large
  multimodal models.
\newblock \emph{arXiv preprint arXiv:2502.09696}, 2025.

\bibitem[Schaul et~al.(2015)Schaul, Quan, Antonoglou, and
  Silver]{schaul2015prioritized}
Tom Schaul, John Quan, Ioannis Antonoglou, and David Silver.
\newblock Prioritized experience replay.
\newblock \emph{arXiv preprint arXiv:1511.05952}, 2015.

\bibitem[Shao et~al.(2024)Shao, Wang, Zhu, Xu, Song, Bi, Zhang, Zhang, Li, Wu,
  Guo, et~al.]{shao2024deepseekmath}
Zhihong Shao, Peiyi Wang, Qihao Zhu, Runxin Xu, Junxiao Song, Xiao Bi, Haowei
  Zhang, Mingchuan Zhang, Y.~K. Li, Yang Wu, Daya Guo, et~al.
\newblock Deepseekmath: Pushing the limits of mathematical reasoning in open
  language models.
\newblock \emph{arXiv preprint arXiv:2402.03300}, 2024.

\bibitem[Toneva et~al.(2019)Toneva, Sordoni, Tachet~des Combes, Trischler,
  Bengio, and Gordon]{toneva2019empirical}
Mariya Toneva, Alessandro Sordoni, R{\'e}mi Tachet~des Combes, Adam Trischler,
  Yoshua Bengio, and Geoffrey~J. Gordon.
\newblock An empirical study of example forgetting during deep neural network
  learning.
\newblock In \emph{International Conference on Learning Representations}, 2019.

\bibitem[Wang et~al.(2024)Wang, Pan, Shi, Lu, Ren, Zhou, Zhan, and
  Li]{wang2024measuring}
Ke~Wang, Junting Pan, Weikang Shi, Zimu Lu, Houxing Ren, Aojun Zhou, Mingjie
  Zhan, and Hongsheng Li.
\newblock Measuring multimodal mathematical reasoning with math-vision dataset.
\newblock \emph{Advances in Neural Information Processing Systems},
  37:\penalty0 95095--95169, 2024.

\bibitem[Yang et~al.(2024)Yang, Zhang, Hui, Gao, Yu, Li, Liu, Tu, Zhou, Lin,
  Lu, Xue, Lin, Liu, Ren, and
  Zhang]{yang2024qwen25mathtechnicalreportmathematical}
An~Yang, Beichen Zhang, Binyuan Hui, Bofei Gao, Bowen Yu, Chengpeng Li,
  Dayiheng Liu, Jianhong Tu, Jingren Zhou, Junyang Lin, Keming Lu, Mingfeng
  Xue, Runji Lin, Tianyu Liu, Xingzhang Ren, and Zhenru Zhang.
\newblock Qwen2.5-math technical report: Toward mathematical expert model via
  self-improvement, 2024.
\newblock URL \url{https://arxiv.org/abs/2409.12122}.

\bibitem[Yu et~al.(2025)Yu, Zhang, Zhu, Yuan, Zuo, Yue, Dai, Fan, Liu, Liu,
  et~al.]{yu2025dapo}
Qiying Yu, Zheng Zhang, Ruofei Zhu, Yufeng Yuan, Xiaochen Zuo, Yu~Yue, Weinan
  Dai, Tiantian Fan, Gaohong Liu, Lingjun Liu, et~al.
\newblock {DAPO}: An open-source llm reinforcement learning system at scale.
\newblock \emph{arXiv preprint arXiv:2503.14476}, 2025.

\bibitem[Yue et~al.(2025{\natexlab{a}})Yue, Zheng, Ni, Wang, Zhang, Tong, Sun,
  Yu, Zhang, Sun, et~al.]{yue2025mmmu}
Xiang Yue, Tianyu Zheng, Yuansheng Ni, Yubo Wang, Kai Zhang, Shengbang Tong,
  Yuxuan Sun, Botao Yu, Ge~Zhang, Huan Sun, et~al.
\newblock Mmmu-pro: A more robust multi-discipline multimodal understanding
  benchmark.
\newblock In \emph{Proceedings of the 63rd Annual Meeting of the Association
  for Computational Linguistics (Volume 1: Long Papers)}, pp.\  15134--15186,
  2025{\natexlab{a}}.

\bibitem[Yue et~al.(2025{\natexlab{b}})Yue, Chen, Lu, Zhao, Wang, Song, and
  Huang]{yue2025does}
Yang Yue, Zhiqi Chen, Rui Lu, Andrew Zhao, Zhaokai Wang, Shiji Song, and Gao
  Huang.
\newblock Does reinforcement learning really incentivize reasoning capacity in
  {LLMs} beyond the base model?
\newblock \emph{arXiv preprint arXiv:2504.13837}, 2025{\natexlab{b}}.

\bibitem[Zhan et~al.(2025)Zhan, Li, Wang, Qu, Liu, Shao, Wong, and
  Cheng]{liang2025exgrpo}
Runzhe Zhan, Yafu Li, Zhi Wang, Xiaoye Qu, Dongrui Liu, Jing Shao, Derek~F.
  Wong, and Yu~Cheng.
\newblock {ExGRPO}: Learning to reason from experience, 2025.
\newblock URL \url{https://arxiv.org/abs/2510.02245}.

\bibitem[Zhang et~al.(2025{\natexlab{a}})Zhang, Fu, Zhang, Fu, Wang, Zhang, and
  Zhou]{zhang2025rlep}
Hongzhi Zhang, Jia Fu, Jingyuan Zhang, Kai Fu, Qi~Wang, Fuzheng Zhang, and
  Guorui Zhou.
\newblock {RLEP}: Reinforcement learning with experience replay for llm
  reasoning.
\newblock \emph{arXiv preprint arXiv:2507.07451}, 2025{\natexlab{a}}.

\bibitem[Zhang et~al.(2024)Zhang, Jiang, Zhang, Lin, Guo, Qiu, Zhou, Lu, Chang,
  Qiao, et~al.]{zhang2024mathverse}
Renrui Zhang, Dongzhi Jiang, Yichi Zhang, Haokun Lin, Ziyu Guo, Pengshuo Qiu,
  Aojun Zhou, Pan Lu, Kai-Wei Chang, Yu~Qiao, et~al.
\newblock Mathverse: Does your multi-modal llm truly see the diagrams in visual
  math problems?
\newblock In \emph{European Conference on Computer Vision}, pp.\  169--186.
  Springer, 2024.

\bibitem[Zhang et~al.(2025{\natexlab{b}})Zhang, Yao, Yu, Liu, Yin, Yin, Yun,
  and Li]{zhang2025ar3po}
Yuheng Zhang, Wenlin Yao, Changlong Yu, Yao Liu, Qingyu Yin, Bing Yin, Hyokun
  Yun, and Lihong Li.
\newblock Improving sampling efficiency in {RLVR} through adaptive rollout and
  response reuse.
\newblock \emph{arXiv preprint arXiv:2509.25808}, 2025{\natexlab{b}}.

\bibitem[Zhou et~al.(2025)Zhou, Shu, Zhao, Wu, Liang, Xiao, Qin, Yang, Xiong,
  Zhang, Huang, and Liu]{zhou2025mlvubenchmarkingmultitasklong}
Junjie Zhou, Yan Shu, Bo~Zhao, Boya Wu, Zhengyang Liang, Shitao Xiao, Minghao
  Qin, Xi~Yang, Yongping Xiong, Bo~Zhang, Tiejun Huang, and Zheng Liu.
\newblock Mlvu: Benchmarking multi-task long video understanding, 2025.
\newblock URL \url{https://arxiv.org/abs/2406.04264}.

\end{thebibliography}
